\documentclass{article}

\usepackage{amsmath}
 \usepackage[preprint]{neurips_2025}

\usepackage[utf8]{inputenc} 
\usepackage[T1]{fontenc}    
\usepackage{hyperref}       
\usepackage{url}            
\usepackage{booktabs}       
\usepackage{amsfonts}       
\usepackage{nicefrac}       
\usepackage{microtype}      
\usepackage{xcolor}         
\usepackage{graphicx}

\usepackage{tabularray}
\usepackage{float}
\usepackage{multirow}
\usepackage{makecell}
\usepackage{subcaption}
\usepackage{mdframed}
\usepackage{CJKutf8}
\usepackage{marvosym}

\title{InteractiveOmni: A Unified Omni-modal Model for Audio-Visual
Multi-turn Dialogue}

\author{%
\vspace{-15px}\\
\scalebox{0.95}{
\textbf{
 Wenwen Tong$^{*}$,
 Hewei Guo$^{*}$,
 Dongchuan Ran$^{*}$,
 Jiangnan Chen$^{*}$,
 Jiefan Lu$^{*}$,
 Kaibin Wang$^{*}$, 
}
}\\
\scalebox{0.95}{
\textbf{
  Keqiang Li$^{*}$,
 Xiaoxu Zhu$^{*}$,
 Jiakui Li$^{*}$,
 Kehan Li,
 Xueheng Li,
 Lumin Li,
 Chenxu Guo,
 }
 }\\
\scalebox{0.95}{
\textbf{
Jiasheng Zhou,
  Jiandong Chen,
  Xianye Wu,
  Jiahao Wang,
  Silei Wu,
  Lei Chen,
  Hanming Deng,
 }
 }\\
 \scalebox{0.95}{
\textbf{
 Yuxuan Song,
 Dinghao Zhou,
 Guiping Zhong,
 Ken Zheng,
 Shiyin Kang\textsuperscript{\Letter},
Lewei Lu\textsuperscript{\Letter}
 }
 }
\vspace{4px}\\
SenseTime Research
 \vspace{4px}\\
\footnotesize{
*~Equal Contribution \;
\Letter~Corresponding Author \;
}
\vspace{4px}\\
\small \url{https://github.com/OpenSenseNova/InteractiveOmni}
}

\begin{document}

\maketitle

\begin{abstract}
We introduce InteractiveOmni, a unified and open-source omni-modal large language model for audio-visual multi-turn interaction, ranging from 4B to 8B parameters, designed to lead the field of lightweight models by offering comprehensive omni-modal understanding and speech generation capabilities.
To achieve this, we integrate the vision encoder, audio encoder, large language model, and speech decoder into a unified model for understanding and generation tasks.
We design a multi-stage training strategy to ensure robust cross-modal capabilities, including pre-training for omni-modal understanding, followed by post-training with speech conversation and audio-visual interaction.
To enable human-like long-term conversational ability, we meticulously curate a multi-turn training dataset that enhances the model's ability to handle complex and multi-turn interactions.
To effectively evaluate the multi-turn memory and speech interaction capabilities, we construct the multi-modal multi-turn memory benchmark and the multi-turn speech interaction benchmark. 
Experiments demonstrate that InteractiveOmni significantly outperforms leading open-source models and provides a more intelligent multi-turn audio-visual experience, particularly in its long-term memory capabilities.
Notably, InteractiveOmni-4B is comparable to the much larger model like Qwen2.5-Omni-7B on general benchmarks, and it can retain 97\% of the performance of the InteractiveOmni-8B while utilizing only 50\% of the model size.
Achieving state-of-the-art results against similarly sized models across image, audio, video understanding, and speech generation tasks, InteractiveOmni is an accessible, open-source foundation for next-generation intelligent interactive systems.
\end{abstract}

\section{Introduction}
\label{sec:intro}

\begin{figure}[h]
    \centering
    \includegraphics[width=0.75\linewidth]{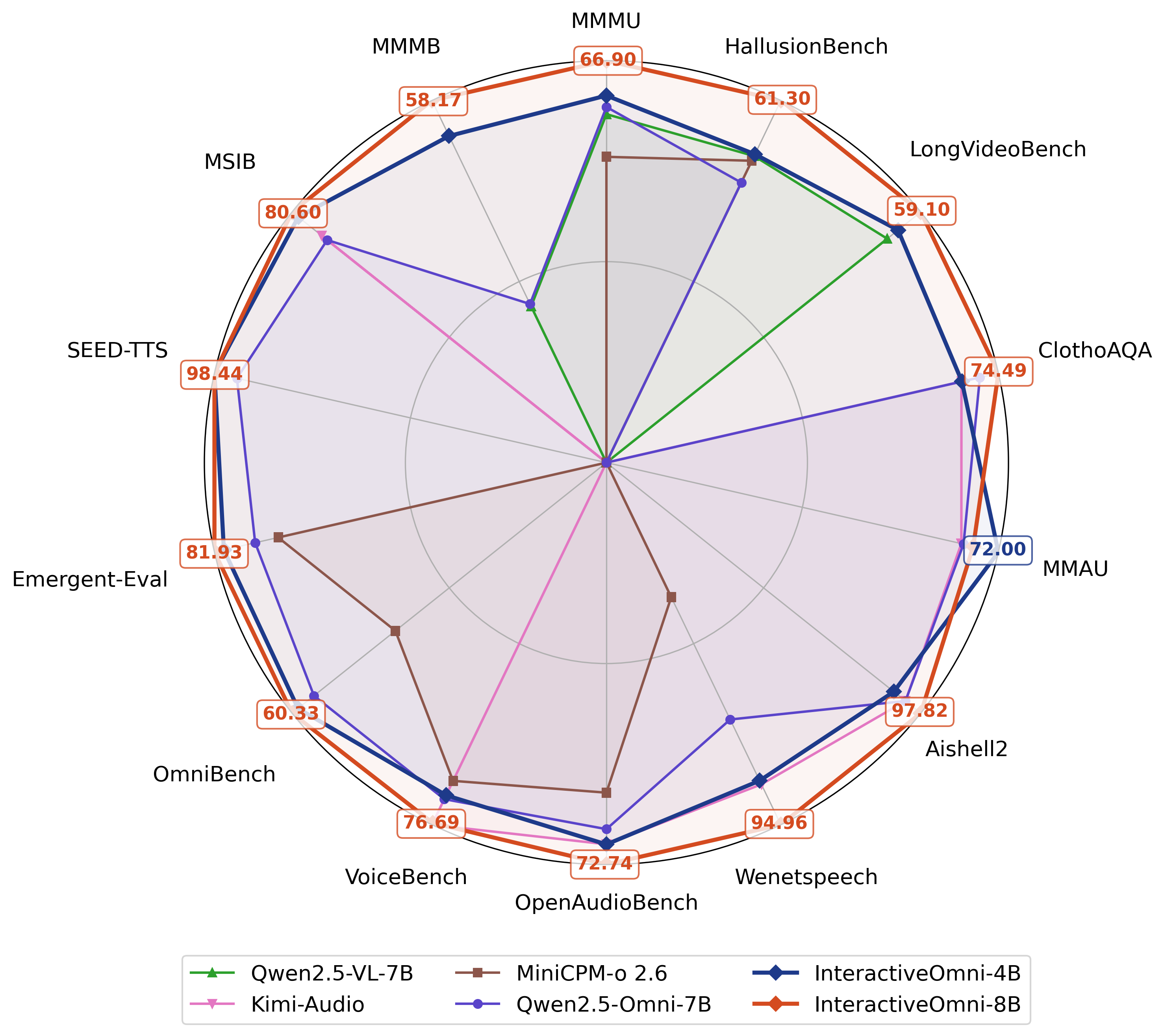}
    \caption{Evaluation across image, video, and audio modalities on open-source benchmarks. InteractiveOmni outperforms the current leading multi-modal models such as Qwen2.5-VL-7B\cite{bai2025qwen25},
    Kimi-Audio \cite{kimiteam2025kimiaudiotechnicalreport},  MiniCPM-o-2.6 \cite{minicpm-v} and Qwen2.5-Omni-7B \cite{xu2025qwen2}.}
    \label{fig:radar_chart}
\end{figure}

\begin{figure}
    \centering
    \includegraphics[width=1\linewidth]{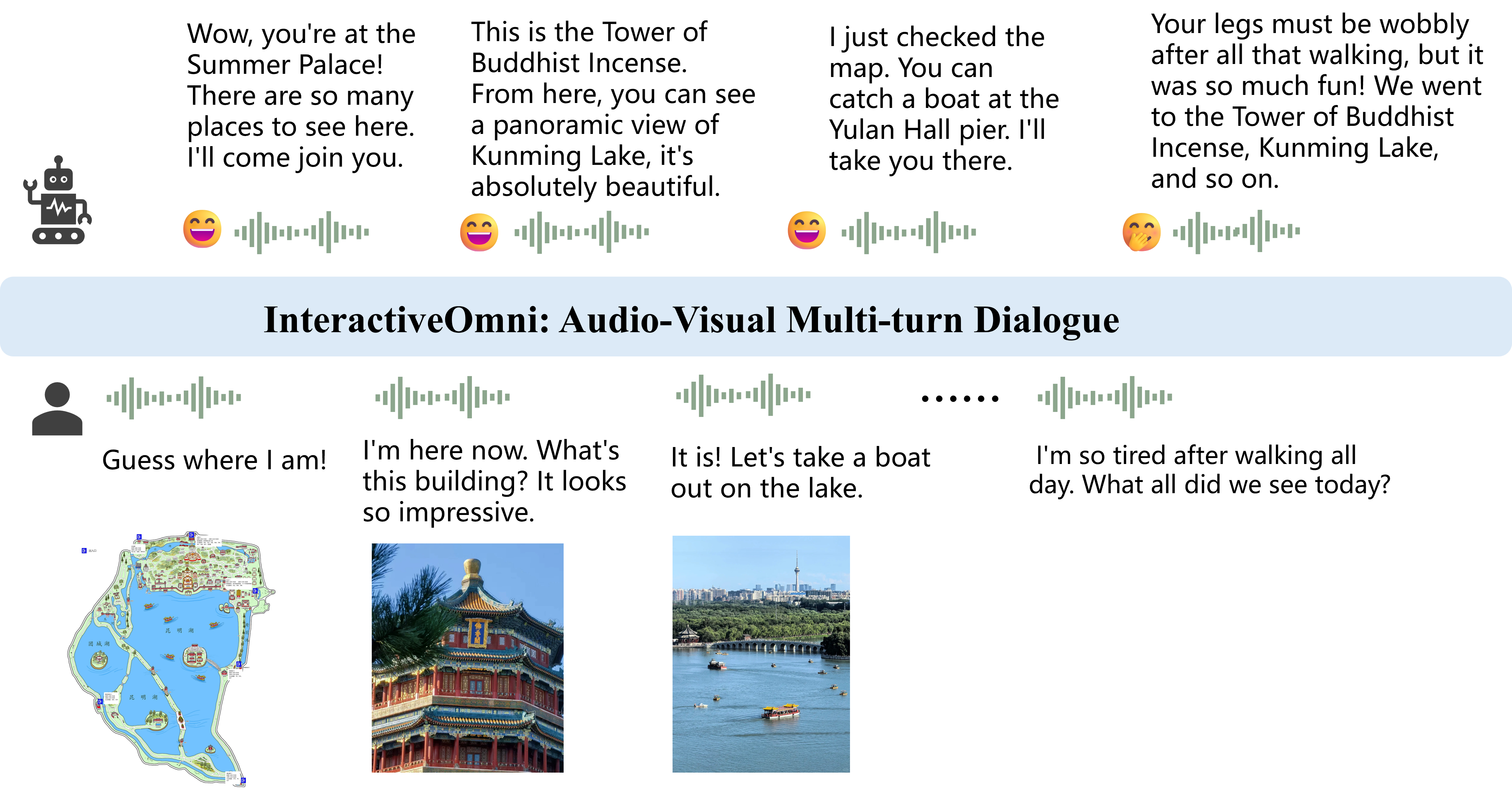}
    \caption{The schematic diagram of multi-turn audio-visual interaction. InteractiveOmni can perceive external audio and video inputs like a human, actively interact with users, and has the capabilities of multi-turn memory and empathy.
}
    \label{fig:demo_interaction}
\end{figure}

Human interaction is fundamentally a holistic and multi-modal experience that integrates sensory information from vision, hearing, and language, supporting natural multi-turn communication and long-term memory, which are the core aspects of intelligence. 
Developing machines with this comprehensive multi-modal multi-turn interactive capability is a critical step toward Artificial General Intelligence (AGI) and represents the next frontier in human-computer interaction \cite{mumuni2025large}. 
Recent breakthroughs in large language models (LLMs) have shown a degree of intelligence, and this is particularly evident in improved problem-solving capabilities and the growing utility in the real world \cite{gpt3,gpt4,guo2025deepseek, yang2025qwen3}.
Furthermore, LLMs can expand their capabilities by integrating vision and audio processing abilities, evolving towards multi-modal large language models (MLLMs), such as vision-language models (VLMs) \cite{llava,bai2025qwen25,internvl,zhu2025internvl3,coreteam2025mimovltechnicalreport,hong2025glm}, audio-language models (ALMs) \cite{chu2024qwen2, kimiteam2025kimiaudiotechnicalreport,wu2025stepaudio2technicalreport}, and omni-modal MLLMs (Omni-MLLMs) \cite{fang2025llamaomni2llmbasedrealtimespoken,li2025baichuan, xu2025qwen2,comanici2025gemini, jiang2024specific,xie2024miniomni2opensourcegpt4ovision}.
Although these works have explored multi-modal capabilities, they mainly focus on understanding ability and single-turn interaction \cite{zhu2025internvl3,kimiteam2025kimiaudiotechnicalreport,xu2025qwen2}, which is different from human-like multi-turn interaction with long-term memory, failing to provide a seamless and integrated user experience on complex, multi-modal interactive tasks in the real world.
Therefore, it is necessary to develop an end-to-end Omni-MLLM capable of understanding omni-modal inputs and synthesizing speech as a response with multi-turn conversational ability, which will serve as the core engine for building the next generation of intelligent interactive experiences and breaking down the barriers between modalities. 
As illustrated in Figure~\ref{fig:demo_interaction}, an Omni-MLLM can serve as an intelligent assistant, offering multi-turn memory and interaction capabilities to accompany us on our travels.

Developing the Omni-MLLM with comprehensive multi-modal interactive capability presents several challenges. 
First, multi-modal alignment is a core difficulty for the development of MLLMs, which has been extensively investigated in VLMs and ALMs \cite{qwenvl, internvl, internvl1.5}. Effectively combining information from heterogeneous data sources, such as images, audio, text, and video, and achieving deep alignment is crucial and much more complex for the training of Omni-MLLM \cite{xie2024miniomni2opensourcegpt4ovision,xu2025qwen2}. 
Second, it is exceptionally challenging to construct an end-to-end unified understanding and generation framework which can process any combination of modal inputs and synchronously generate streaming text and audio \cite{ai2025ming, xu2025qwen2}.
Finally, enhancing the model's strong interactive capabilities and speech emotional expressiveness is central to its ultimate practical value  \cite{geng2025osum,higgsaudio2025,peng2025vibevoice}, including the long-term memory, human-like emotion and empathy, and the maintenance of contextual consistency and logical coherence in the multi-turn dialogues. Current MLLMs exhibit limited capabilities for real-world interaction, and there is also a lack of benchmarks to evaluate the multi-turn interaction ability and practicality \cite{sirdeshmukh2025multichallenge}.

To address these challenges, we propose InteractiveOmni, an Omni-MLLM with end-to-end understanding and generation capabilities, providing intelligent multi-turn interactive experience.
We employ a single architecture to process and generate data across all modalities, achieving an end-to-end workflow from omni-modal input to text and speech output. To address the omni-modal alignment problem, we propose the omni-modal pre-training and post-training strategy. Through meticulously designed pre-training tasks, the model learns the intrinsic correlations between different modalities at an early stage. Subsequently, during the post-training phase, we leverage the instruction tuning and direct preference optimization (DPO) to further strengthen the cross-modal capabilities. Furthermore, to enhance the interactive experience, we constructe highly interactive multi-turn data combined with post-training for optimization, focusing on improving the model's performance in memory, empathy, and contextual understanding to make its interactions more human-like and intelligent. 
To effectively evaluate multi-turn memory and speech interaction, we meticulously construct new benchmarks: the multi-modal multi-turn memory benchmark (MMMB) and the multi-turn speech interaction benchmark (MSIB), to address the shortcomings of existing multi-turn dialogue benchmarks.

We develop InteractiveOmni based on open-source models \cite{yang2025qwen3, internvl, du2024cosyvoice, whisper}, achieving comprehensive leading performance in multi-modal understanding and generation tasks. Specifically, in visual understanding tasks, InteractiveOmni is comparable to state-of-the-art vision-language models such as Qwen2.5-VL-7B \cite{bai2025qwen25} and InternVL3.5-8B\cite{wang2025internvl35advancingopensourcemultimodal}. 
For audio understanding and speech conversation tasks, InteractiveOmni's performance rivals leading audio-language models, including Kimi-Audio \cite{kimiteam2025kimiaudiotechnicalreport} and Step-Audio-Chat \cite{wu2025stepaudio2technicalreport}. 
Furthermore, InteractiveOmni delivers superior performance on omni-modal benchmarks, outperforming models like MiniCPM-o-2.6 \cite{minicpm-v}, Qwen2.5-Omni-7B \cite{xu2025qwen2}, and Ming-Lite-Omni \cite{ai2025ming}.
In addition, InteractiveOmni demonstrates superior performance in the comprehensive multi-turn benchmarks, showcasing its excellent interactive capabilities in real-world applications.
The key contributions of InteractiveOmni can be summarized as follows:
\begin{itemize}
    \item We propose a unified omni-modal model that can simultaneously receive inputs such as images, audio, text, and video and directly generate coherent text and speech streams, achieving truly integrated multi-turn interaction. 
    \item InteractiveOmni achieves state-of-the-art performance against similarly sized multi-modal large language models on several mainstream open-source benchmarks for image, audio, and video understanding, as well as speech conversation. Notably, InteractiveOmni-4B is comparable to the much larger Qwen2.5-Omni-7B on various  benchmarks.
    \item  InteractiveOmni demonstrates excellent interactive performance with multi-turn and long-term memory capabilities. To effectively evaluate this capability, we construct the multi-turn benchmarks such as MMMB and MSIB, specifically for assessing the multi-turn, multi-modal interactive capabilities.

\end{itemize}
\section{Method}
\label{sec:method}

\subsection{Architecture}
\begin{figure}[t]
    \centering
    \includegraphics[width=0.9\linewidth]{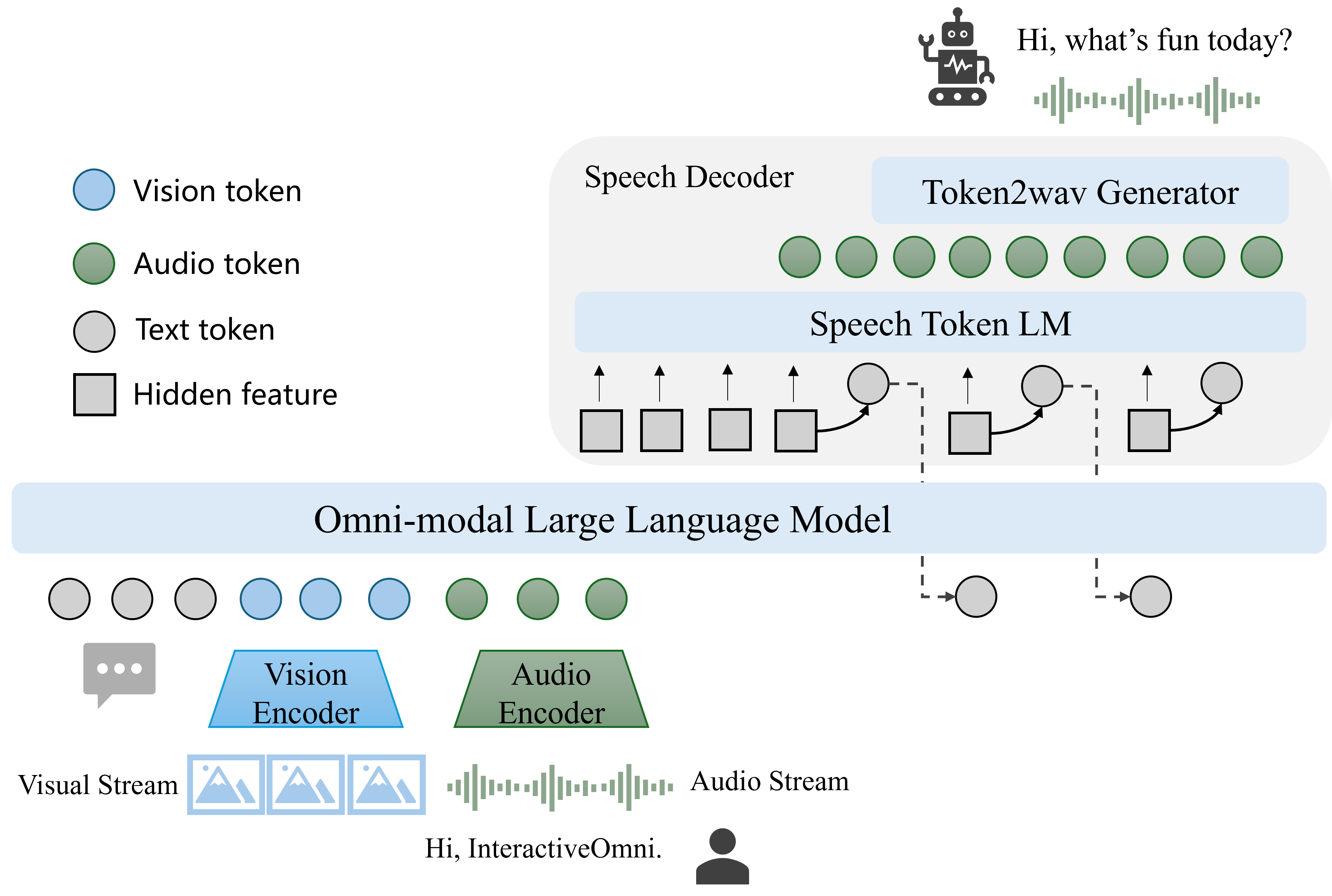}
    \caption{The overview framework of InteractiveOmni. InteractiveOmni is composed of vision encoder, audio encoder, LLM decoder and streaming speech decoder. The extracted visual and audio tokens are processed by the LLM to generate text tokens and speech tokens sequentially. }
    \label{fig:model_architecture}
\end{figure}

As shown in Figure~\ref{fig:model_architecture}, InteractiveOmni is a unified model capable of perceiving omni-modal inputs such as image, video, audio, and text, while generating text and speech sequentially, achieving end-to-end omni-modal perception and generation. InteractiveOmni consists of vision encoder, audio encoder, LLM decoder, speech-token LM, and token2wav speech generator. The architecture of InteractiveOmni-4B and InteractiveOmni-8B is shown in Table~\ref{tab:architecture_sensenova_omni}.

\begin{table}[htbp]
  \centering
  \caption{The architecture of InteractiveOmni models.}
  \label{tab:architecture_sensenova_omni}
  \begin{tabular}{lcccc}
    \toprule
    \textbf{Module} & \textbf{Vision Encoder} & \textbf{Audio Encoder} & \textbf{LLM} & \textbf{Speech Decoder}\\
    \midrule
    InteractiveOmni-4B &  InternViT & Whisper &  Qwen3-4B & Cosyvoice2\\
    InteractiveOmni-8B &  InternViT & Whisper &  Qwen3-8B & Cosyvoice2\\
    \bottomrule
  \end{tabular}
\end{table}

We adopt the audio encoder from the Whisper-large-v3 model \cite{whisper} due to its strong performance on audio understanding tasks.
Similar to the preprocessing of audio in Qwen2-Audio \cite{chu2024qwen2}, we resample the input audio data to a frequency of 16kHz and convert the raw waveform into the 128-channel mel-spectrogram. In addition, we add a pooling layer to downsample the output length of audio to the frame rate of 25Hz, meaning that one second of audio is represented by 25 tokens. An audio adapter with a two-layer MLP projector is employed to connect the audio encoder to LLM.

We utilize the InternViT-300M \cite{internvl2.5} as the vision encoder to handle the image and video inputs. In terms of the data preprocessing, we employ the dynamic resolution strategy to divide the images into tiles of 448x448 pixels based on the resolution and aspect ratio of the image \cite{internvl1.5, internvl, zhu2025internvl3}. Since the representation of high-resolution image and long video inputs requires a large number of visual tokens, we employ the pixel shuffle operation to reduce the number of visual tokens to one-sixteenth of their original number. Thus, a 448x448 image is represented by 64 visual tokens in our model. Finally, a two-layer MLP projector is utilized to map the visual features into the embedding space of the LLM.

We use the pretrained Qwen3 \cite{yang2025qwen3} as the LLM decoder, considering its outstanding performance on various text benchmarks.
The LLM takes the visual features and audio features as input, and decodes text tokens sequentially.
Our speech decoder, based on Cosyvoice2 \cite{du2024cosyvoice}, consists of a speech token LM and a token2wav speech generator.
To generate the speech in a streaming fashion, we interleave the generated text tokens and speech tokens in a 5:25 ratio. Specifically, for every five text tokens generated, we pass the text token embeddings and corresponding hidden features to the speech token LM to generate 25 speech tokens, ensuring efficient and seamless speech synthesis. Then, the 25 speech tokens are passed to the token2wav generator to produce the final speech output. For the speech-to-speech conversational scenario, the generated speech style can also be controlled by the user instruction and text hidden features to generate more emotionally expressive speech.

InteractiveOmni is fully trained end-to-end for the omni-modal understanding and generation task based on the hidden embeddings connecting the LLM, vision encoder, audio encoder, and speech decoder. The training datasets are explained in detail in Section~\ref{sec:datasets}, and the training procedure of InteractiveOmni is given in Section~\ref{sec:training}.

\subsection{Datasets}\label{sec:datasets}

\begin{figure}[h]
    \centering
    \includegraphics[width=0.95\linewidth]{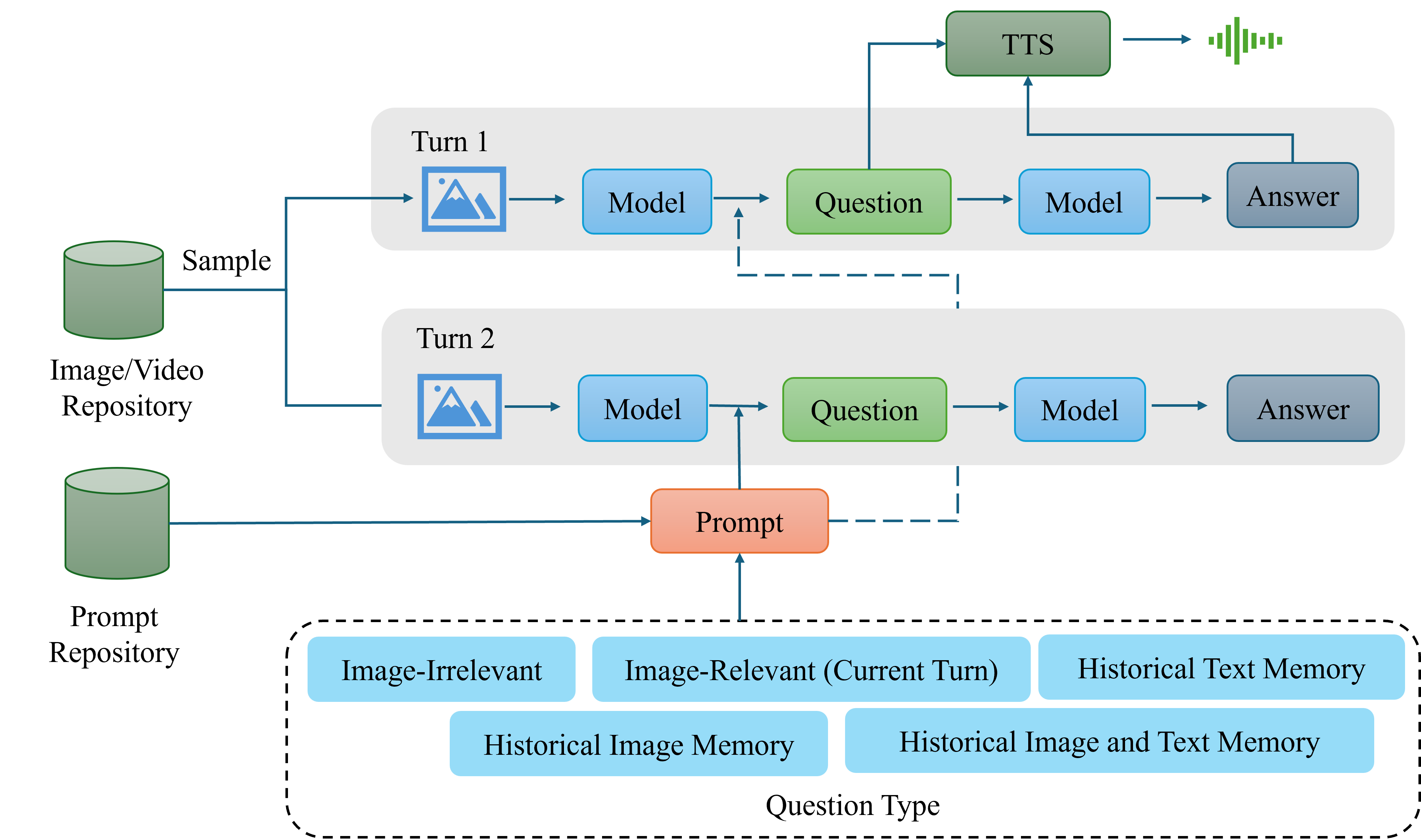}
    \caption{Data construction pipeline for multi-turn dialogue. In each turn, the visual element is sampled from a dedicated image and video repository. The corresponding question is then generated by a vision-language model using a specific prompt tailored to the desired question type. To ensure the dialogue effectively tests long-term memory, we specifically design turns that require recalling historical images and previous dialogue text. Finally, the generated text-format question and answer can be transformed into speech-based question-answer pairs using a TTS system, facilitating end-to-end training.}
    \label{fig:data_pipeline}
\end{figure}

To enhance the performance of audio-visual multi-turn dialogue and improve the long-term memory capacity, we have carefully constructed a multi-turn data generation pipeline. As illustrated in Figure~\ref{fig:data_pipeline}, we first establish a comprehensive repository of images and videos. For each dialogue turn, the visual element is sampled from this repository to serve as the visual input. 
The corresponding question is then generated by a vision-language model using a specific prompt tailored to the desired question type. 
The questions in each turn can be categorized based on the scope of the information required for a correct answer. Specifically, the questions can be categorized into five types:

\begin{itemize}
    \item \textbf{Image-Irrelevant}: The question is a pure text-based query that is completely independent of the current image and the dialogue history. 
    \item \textbf{Image-Relevant (Current Turn)}: The question is visually-grounded and can be answered solely by analyzing the current image and the text of the current question.
    \item \textbf{Historical Image Memory}: The question requires the model to recall and reason about information presented in a previously shown image within the dialogue history. 
    \item \textbf{Historical Text Memory}: The question is grounded in the previous turns of the dialogue text, but does not require reference to any specific image.
    \item \textbf{Historical Image and Text Memory}: The question necessitates the integration of information from both the historical dialogue text and images.
\end{itemize}
We primarily construct multi-turn dialogue data within 20 turns based on this data pipeline. To facilitate end-to-end training, we can also transform the generated text-format question and answer into speech-based question-answer pairs using the TTS system.

In the subsequent sections, we provide a detailed breakdown of the training data and the open-source data utilized. This includes data categorized by modality and task: image understanding data, video understanding data, audio understanding data, omni-modal understanding data, audio generation data, and end-to-end dialogue data.

\subsubsection{Image Data}\label{sec:image_data}
To enhance the visual understanding capabilities of the model, we curate a comprehensive collection of multi-modal datasets with approximately 12 million image-text pairs for post-training, including open-source, synthetic, and proprietary in-house data. This visual corpus encompasses multiple domains, such as general question answering (GeneralQA), optical character recognition (OCR), document understanding, mathematics, science, knowledge, and perception. For a detailed statistical breakdown of the open-source dataset's composition, please refer to Table \ref{table:visual_data_statistics}.

\begin{table}[ht]
\centering
\footnotesize
\caption{Detailed statistics of the training data of open-source image data.}
\begin{tabular}{@{}c p{10cm}}
\toprule
\textbf{Task} & \textbf{Datasets}
\\ \midrule
\multirow{2}{*}{OCR}   & TextVQA\cite{textvqa}, OCRVQA\cite{ocrvqa}, ST-VQA\cite{stvqa_biten2019scenetextvisualquestion}, LSVT\cite{sun2019LSVT_icdar2019competitionlargescale}, ArT\cite{chng2019ART_icdar2019robustreadingchallenge}, CTW\cite{yuan2019ctw}, RCTW\cite{rctw_shi2017icdar2017}, COCO-Text\cite{veit2016cocotextdatasetbenchmarktext}, MTVQA\cite{tang2024mtvqa}, ReCTs\cite{liu2019ReCTs_icdar2019robustreading}, MathWriting\cite{gervais2025mathwritingdatasethandwrittenmathematical}
\\ \midrule
\multirow{2}{*}{Document Understanding}    & InfographicVQA\cite{mathew2022infographicvqa}, LLaVAR\cite{zhang2023llavar}, FigureQA\cite{kahou2017figureqa}, MapQA\cite{chang2022mapqadatasetquestionanswering}, SROIE\cite{sroie_huang2019icdar2019}, Docmatix\cite{docmatix_laurençon2024building}, DocVQA\cite{docvqa}
\\ \midrule
\multirow{2}{*}{GeneralQA}  & VQAv2\cite{VQAv2}, Visual7W\cite{zhu2016visual7wgroundedquestionanswering}, ViRL39K\cite{ViRL39K_vl-rethinker}, MMDU\cite{liu2024mmdumultiturnmultiimagedialog}, VIST\cite{VIST_huang2016visual}, GQA\cite{gqa}, OKVQA\cite{marino2019ok_okvqa}
\\ \midrule
\multirow{1}{*}{Science}   & TQA\cite{TQA_Kembhavi_2017_CVPR}, AI2D\cite{kembhavi2016diagram}, ScienceQA\cite{lu2022learn_scienceqa}
\\ \midrule
\multirow{1}{*}{Mathematics}  & GeoQA+\cite{chen2021geoqa}, Geometry3K\cite{lu2021inter}, MathQA\cite{yu2024MATHQA_metamathbootstrapmathematicalquestions}, MAVIS\cite{zhang2024mavismathematicalvisualinstruction}, UniGeo\cite{chen2022unigeounifyinggeometrylogical}
\\ \midrule
\multirow{1}{*}{Knowledge}   & A-OKVQA\cite{schwenk2022aokvqabenchmarkvisualquestion}, ART500K\cite{art500k_mao2019visual}, ViQuAE\cite{lerner2022viquae}, KVQA\cite{shah2019kvqa}
\\ \midrule
\multirow{2}{*}{Perception}  & PuzzleVQA\cite{chia2024puzzlevqadiagnosingmultimodalreasoning}, Spot-the-diff\cite{zou2022spotthedifferenceselfsupervisedpretraininganomaly}, VSR\cite{liu2023VSR_visualspatialreasoning}, TallyQA\cite{acharya2019tallyqa}, IconQA\cite{lu2021iconqa}, RefCOCO\cite{refcoco}, Object365\cite{shao2019Object365}
\\ \bottomrule
\end{tabular}
\label{table:visual_data_statistics}
\end{table}

\subsubsection{Video Data}\label{sec:video_data}
The video data is composed of various data with 5 million video-text pairs covering several distinct tasks such as the short caption, detailed caption, video question-answering (VideoQA) and Video Temporal Grounding (VTG). This strategic composition of the dataset ensures that the model's performance can be thoroughly improved across a variety of complexities, from high-level summarization to fine-grained temporal and semantic understanding. A detailed breakdown of the open-source dataset composition is provided in Table \ref{tab:video_data}.

\begin{table}[ht]
\centering
\footnotesize
\caption{Detailed statistics of the training data of open-source video data.}
\begin{tabular}{@{}c p{10cm}}
\toprule
\textbf{Task} & \textbf{Datasets}
\\ \midrule
Short Caption  & InternVid-10M\cite{wang2023internvid}, WebVid\cite{Bain21webvid}, OpenVid\cite{nan2024openvid}, TextVR\cite{wu2025largeTextVR}, Mementos\cite{ransford2011mementos}
\\ \midrule
\multirow{2}{*}{Detailed Caption}   & ShareGPT4Video\cite{chen2024sharegpt4video}, Vript\cite{yang2024vript}, LSMDC\cite{rohrbach2015dataset}, Mementos\cite{ransford2011mementos}, PE-Video\cite{cho2025perceptionlm}, LLaVA-Video\cite{zhang2025llavavideovideoinstructiontuning}
\\ \midrule
\multirow{2}{*}{VideoQA}  & STAR\cite{xie2025starspatialtemporalaugmentationtexttovideo}, EgoTaskQA\cite{jia2022egotaskqa}, TVQA\cite{lei2018tvqa}, HiREST\cite{zala2023hierarchical}, PerceptionTest\cite{patraucean2023perception}, VideoGPT+\cite{Maaz2024VideoGPT+}, CLEVRER\cite{yi2019clevrer}
\\ \midrule
\makecell[c]{VTG}   & ET-Instruct-164k\cite{liu2024etbench}, hdvila\cite{xue2022advancing}, Koala-36M\cite{wang2025koala}, HiREST\cite{zala2023hierarchical}
\\ \bottomrule
\end{tabular}
\label{tab:video_data}
\end{table}



\subsubsection{Audio Data}
The audio understanding data is built on a massive dataset of over 240,000 hours, including speech, sound, and music data, as shown in Table~\ref{tab:stage2_data_summary}. The primary component is dedicated to automatic speech recognition (ASR), which comprises over 187,000 hours of English and Chinese speech. The ASR data is sourced from academic benchmarks, crowdsourcing, and in-house collections, constituting approximately 76\% of our entire audio dataset and nearly 90\% of all speech-related data, providing a robust foundation for speech understanding.

To achieve a more comprehensive and nuanced understanding of audio, the remaining portion of audio data is strategically allocated to a variety of specialized tasks as indicated in Table~\ref{tab:stage2_data_summary}.  For speech-related applications, this includes a substantial corpus of over 10,000 hours, such as translation, speech question answering, and emotion recognition. Beyond speech, we incorporate over 18,000 hours of general sound data, with the majority dedicated to question answering and captioning tasks. Furthermore, over 16,000 hours of music data are used for training music-based question answering, enabling the model to interpret a wide spectrum of complex audio signals.

\begin{table}[ht]
\centering
\footnotesize
\caption{Summary of datasets for audio understanding tasks, including speech, sound, and music.}
\label{tab:stage2_data_summary}
\begin{tblr}{
  width = \linewidth,
  colspec = {Q[l,m] Q[c,m] X[j] Q[c,m]},
  row{1} = {c, font=\bfseries},
  cell{2}{1} = {r=5}{c,m},
  cell{7}{1} = {r=3}{c,m},
  cell{10}{1} = {c,m},
  hline{1,11} = {-}{0.08em},
  hline{2,7,10} = {-}{0.05em},
  column{2} = {c,m}, 
}
Category & Task & Datasets & Hours
\\
\textbf{Speech} &
ASR &
AISHELL~\cite{bu2017aishell,du2018aishell2,shi2020aishell3}, ChildMandarin~\cite{zhou2024childmandarin}, CommonVoice~\cite{ardila2019commonvoice}, Emilia~\cite{he2024emilia}, Fleurs~\cite{conneau2023fleurs},  GigaSpeech~\cite{chen2021gigaspeech}, Libriheavy~\cite{kang2024libriheavy}, LibriSpeech~\cite{panayotov2015librispeech}, LibriTTS~\cite{zen2019libritts}, MLS-ENG~\cite{Pratap2020MLSAL}, SPGISpeech~\cite{o2021spgispeech}, WenetSpeech~\cite{zhang2022wenetspeech} &
187,942
\\
&
QA &
Open-ASQA-Speech~\cite{gong2023joint}, SLURP~\cite{bastianelli2020slurp} &
2,444
\\
&
Emotion Recognition &
MELD~\cite{poria2018meld}, IEMCAP~\cite{busso2008iemocap}, CSEMOTIONS~\cite{tian2025marco}, Nonverbal~\cite{borisov2025nonverbaltts} &
38
\\
&
Translation &
CoVoST-1~\cite{wang-etal-2020-covost}, CoVoST-2~\cite{wang2020covost},GigaST~\cite{ye2022gigast} &
10,277
\\
&
Inhouse &
- &
11,282
\\
\textbf{Sound} &
QA &
Clotho-AQA~\cite{lipping2022clothoaqa}, CompA-R-Instructions~\cite{ghosh2024gama}, Open-ASQA-Non-Speech~\cite{gong2023joint}, VocalSound~\cite{gong2022vocalsound} &
10,997
\\
&
Sound Classification &
Cochlscene~\cite{jeong2022cochlscene}, ESC-50~\cite{piczak2015esc}, MACS~\cite{morato2021diversity}, TAU~\cite{wang2021curated},  Urbansound8k~\cite{Salamon:UrbanSound:ACMMM:14}, VggSound~\cite{chen2020vggsound} &
835
\\
&
Caption &
AudioCaps~\cite{kim2019audiocaps}, Auto-ACD~\cite{sun2024auto}, Clotho~\cite{drossos2020clotho}, SoundDescs~\cite{koepke2022audio}, Epidemic Sounds~\cite{huh2025epic}, Wavcaps~\cite{mei2024wavcaps}, WavText5k~\cite{deshmukh2022audio} &
6,397
\\
\textbf{Music} &
QA &
FMA~\cite{fma_dataset}, MagnaTagATune~\cite{wolff2012systematic}, FSD2018~\cite{fonseca2018general}, MusicBench~\cite{melechovsky2023mustango},  MusicQA~\cite{liu2024music} &
16,605
\\
\end{tblr}
\label{audio datasets}
\end{table}

\subsubsection{Omni-modal Data}\label{sec:omni_modal_data}
We curate a comprehensive omni-modal dataset by integrating open-source, synthetic, and in-house annotated data. Spanning multiple tasks and modalities, the dataset comprises approximately 15 million data pairs, including audio-to-text, image-audio-to-text, and video-audio-to-text combinations.
Based on the image data in Section~\ref{sec:image_data} and video data in Section~\ref{sec:video_data}, we curate a large volume of speech interaction data by converting the text-based question from the multi-modal dataset into speech-based question using the TTS system.
Furthermore, to enhance the ability for spoken dialogue with speech input and text response, we construct a dataset of spoken dialogue through rewriting original formatted multi-modal data or using the LLM to generate spoken conversational data.
We also develop a sophisticated, multi-stage data processing pipeline to synthesize multi-turn image, audio, and text conversational data as shown in Figure~\ref{fig:data_pipeline}, significantly improving the model's long-term memory and multi-turn interactive capabilities.
Therefore, the model demonstrates comprehensive and omni-modal understanding capability, which is enabled by the integration of this high-quality omni-modal data.

\subsubsection{Text-to-Speech Dataset}
This category of data mainly consists of two parts, including the basic speech synthesis data and style-controllable speech synthesis data, as detailed in Table~\ref{tab:speech_task_duration}. 
The basic synthesis data consists of approximately 202,000 hours of large-scale public Chinese-English text-speech paired corpora. This dataset is crucial for supporting fundamental speech synthesis capabilities, including linguistic coverage, cross-domain robustness, and generalization.
Additionally, we construct about 1,000 hours of style-controllable data, which is guided by natural language instructions along four dimensions: speech rate, emotional tone, dialect, and character persona. 
These instructions were generated by Qwen3 \cite{yang2025qwen3} and subsequently converted into high-quality speech using our in-house TTS system.

\begin{table}[htbp]
  \centering
  \caption{Detailed statistics of training data for the speech generation task.}
  \label{tab:speech_task_duration}
  \begin{tabular}{lc}
    \toprule
    \textbf{Task} & \textbf{Hours} \\
    \midrule
    Speech Synthesis & 202k \\
    Style-controllable Speech Synthesis & 1k \\
    Speech-to-Speech Chat & 11k \\
    Style-controllable Speech-to-Speech Chat & 11k \\
    \bottomrule
  \end{tabular}
\end{table}

\subsubsection{Speech-to-Speech Dataset}
The speech-to-speech dataset supports end-to-end model training by enabling the system to comprehend user speech inputs and generate contextually appropriate spoken responses.
We curate a large volume of speech-to-speech data by converting the text-based question-answering into speech-based question-answering using the TTS system based on the omni-modal data in Section~\ref{sec:omni_modal_data}. In addition, we construct colloquial multi-turn conversational data to improve the naturalness of human-machine interaction. 
This process yields approximately 11,000 hours of speech conversation data. 
Furthermore, we develop the data pipeline to generate style-controllable speech-to-speech dialogue data of approximately 11,000 hours, covering different speaking styles such as emotion, speech rate, and role-play.
A detailed breakdown of the speech-to-speech dialogue data is given in Table~\ref{tab:speech_task_duration}. 
As a result, the model can produce highly expressive and human-like speech responses for the speech-to-speech question-answering tasks.

\subsection{Training}\label{sec:training}
The training process of InteractiveOmni comprises two main stages. In the first stage, we perform omni-modal pre-training to achieve alignment across audio, image, video, and text modalities. The second stage involves post-training, which enhances the model's ability to follow instructions and engage in audio-visual interactions. A detailed description of the training procedure is provided in Section~\ref{sec:pre_training} and Section~\ref{sec:post_training}, respectively.

\subsubsection{Pre-training}\label{sec:pre_training}
For the initial pre-training stage, InteractiveOmni is initialized with Qwen3 \cite{yang2025qwen3} as the pretrained textual LLM, InternViT-300M \cite{internvl2.5} as the vision encoder, and  Whisper-large-v3 model \cite{whisper} as the audio encoder.
The model is pre-trained on a diverse mixture of datasets, comprising image-text pairs, interleaved image-text data, video-text pairs, audio-text pairs, multimodal question-answering data, and pure text corpora.
The instruction-following data is also incorporated  in the pre-training stage to further improve the model's performance. 
To improve training efficiency, we employ a data-packing strategy, with the maximum token length set to 32,768 to better accommodate long video sequences and multi-turn audio-visual interactions.

The pre-training methodology is structured in three progressive stages to incrementally incorporate additional tasks. 
In the first stage, we leverage vision-text data to train the vision encoder, establishing a foundational alignment between image, video, and text. 
The second stage focuses on the audio encoder, which is trained with audio-text data to align the audio and text modalities. 
The final stage integrates a vast and diverse corpus of mixed multi-modal data, including audio-image, audio-video, audio-image-text, and audio-video-text data, to improve the model’s comprehensive understanding across all modalities. 
Extensive evaluations demonstrate that the resulting pre-trained model exhibits strong performance on a wide range of omni-modal understanding tasks.

\subsubsection{Post-training}\label{sec:post_training}
For the post-training stage, we focus on the improvement of audio-visual interaction and speech-to-speech conversational ability to achieve the end-to-end interaction. 
We conduct multi-task supervised fine-tuning to enhance the model’s ability to follow instructions in audio-visual conversations involving speech-based questions and text-based answers. 
We curate a large volume of audio-visual interaction data by converting text-based questions from a multimodal question-answering dataset into speech format using a TTS system, including the speech-to-text and image-speech-to-text data as shown in Section~\ref{sec:omni_modal_data}. 
In this stage, the audio encoder, vision encoder, and LLM are trainable to improve the model's performance with multi-modal inputs.
The model can acquire the capabilities of audio-visual understanding and dialogue capabilities after this training stage. 
We can then directly integrate an external TTS system to enable full audio-visual conversation in a speech-to-speech format.

To achieve end-to-end dialogue, we integrate the speech decoder into the architecture to enable end-to-end speech conversation as shown in Figure~\ref{fig:model_architecture}, avoiding the need for an external TTS system.
First, we utilize large-scale Chinese–English TTS corpora to train the Speech LM module and adaptor, aligning text tokens with speech tokens.
To support streaming speech output, we interleave the generated text tokens and speech tokens in a 5:25 ratio \cite{du2024cosyvoice2scalablestreaming}. 
To address the abundance of simple samples in the TTS-generated data, we employ a hard sample mining strategy to enhance model robustness and performance. 
Subsequently, the model is trained on speech-to-speech conversational data to enhance end-to-end audio-visual interaction capabilities.
During this phase, we curate high-quality multi-turn speech-to-speech and image-speech-to-speech dialogue data to improve contextual conversational ability as shown in Figure~\ref{fig:data_pipeline}. 
Additionally, style-controllable speech-to-speech data is incorporated to strengthen the emotional expressiveness of the generated speech.


Lastly, we utilize the DPO \cite{rafailov2023direct} to improve the quality of generated content. Experiments show that DPO is effective for the multi-turn conversational scenario, which can enhance the multi-turn interactive experience. Specifically, for the multi-turn conversation, we mainly optimize the final round to improve the interactive experience. 
Furthermore, we find that the model merging technique \cite{li2025model} is effective in enhancing the model performance. We apply this technique in the pre-training stage, merging the checkpoint of the pre-trained model and the continuously trained model to improve the model's performance on the omni-modal understanding tasks.

\section{Evaluation}
\label{sec:exps}

We conduct extensive evaluations of InteractiveOmni on both in-house multi-turn conversational benchmarks and open-source benchmarks, covering the omni-modal understanding and speech generation tasks. 
We compare InteractiveOmni with  proprietary models such as GPT-4o \cite{gpt4o}, Gemini~\cite{comanici2025gemini} and open-source models including MiniCPM-o-2.6~\cite{minicpm-v}, Qwen2.5-Omni-7B \cite{xu2025qwen2}, Kimi-Audio \cite{kimiteam2025kimiaudiotechnicalreport}, Qwen2.5-VL \cite{bai2025qwen25}, and InternVL3 \cite{zhu2025internvl3} across image, video, audio and text benchmarks.

\subsection{Multi-turn Benchmarks}

\subsubsection{Multi-modal Multi-turn Memory Benchmark(MMMB)}

\paragraph{Benchmark Introduction.} We construct the multi-modal multi-turn memory benchmark (MMMB) to evaluate the multi-turn performance of MLLMs owing to the poor performance of multi-turn interaction capability for current MLLMs. The central objective of MMMB is to investigate the following question: How effectively can MLLMs utilize information from historical turns to answer the question related to the historical images and text in the multi-turn interaction? 

MMMB consists of 300 dialogue groups, each with a maximum of 15 turns, designed to assess the multi-turn memory of historical text and images.
In each multi-image multi-turn dialogue, textual and visual information is introduced progressively across turns. 
The final turn poses a question that can only be answered by accurately utilizing information from the historical dialogue context. 
For performance evaluation, we exclusively assess the model’s response in this final turn, treating all preceding turns as contextual history.
Based on the memory information required to answer the final question, the data can be categorized into three types:
\begin{itemize}
    \item \textbf{Text Memory}: Answers derived solely from textual information within the dialogue history.
    \item \textbf{Image Memory}: Answers that rely on images from previous turns.
    \item \textbf{Mixed Memory}: Answers that require both textual and visual information from the history.
\end{itemize}

\begin{figure}[h]
    \centering
    \includegraphics[width=0.95\linewidth]{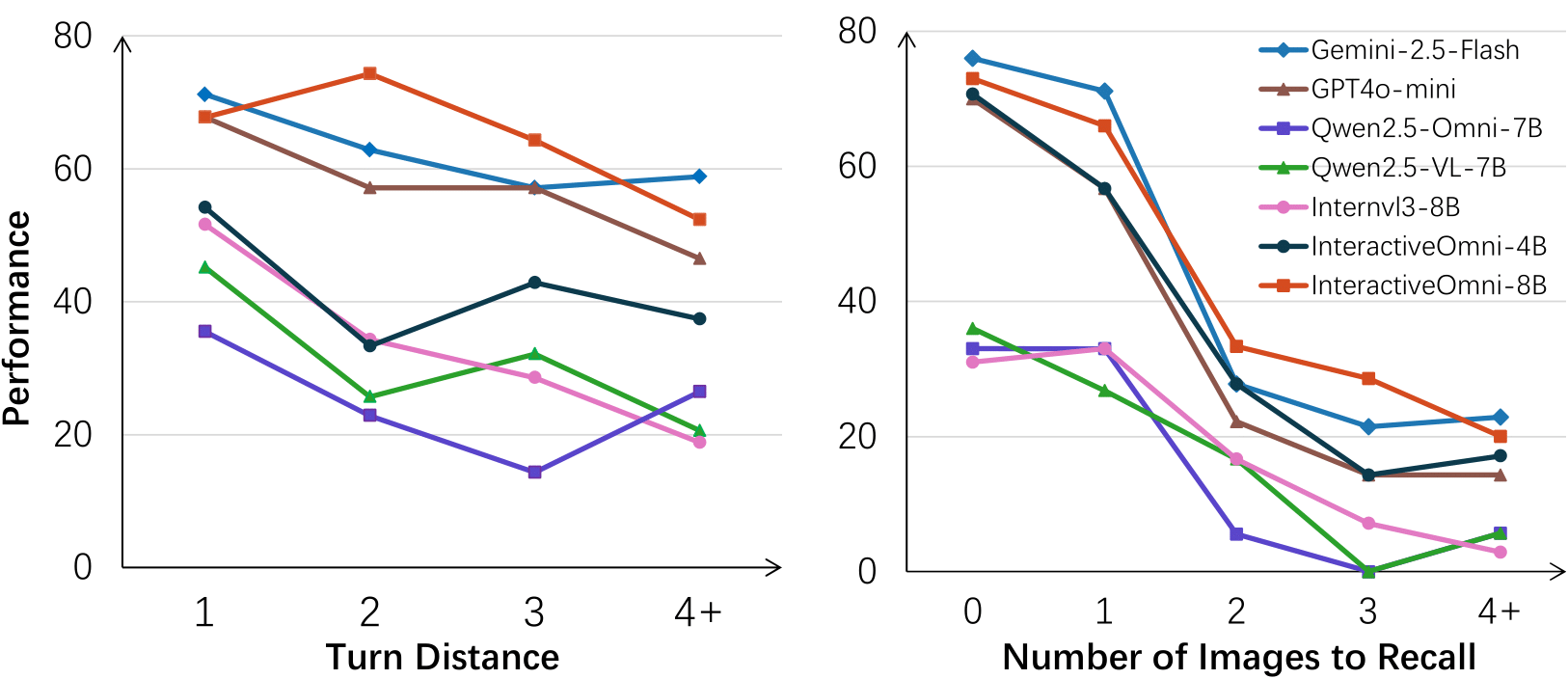}
    \caption{The sketch of performance degradation with the increase of recall burden considering the turn distance and number of memorized images. 
    InteractiveOmni is comparable to proprietary models like GPT-4o-mini and Gemini-2.5-Flash, consistently outperforming open-source models such as InternVL3-8B, Qwen2.5-VL-7B, and Qwen2.5-Omni-7B.
}
    \label{fig:mmmb_performance_degradation}
\end{figure}

\paragraph{Evaluation Results.} We evaluate InteractiveOmni against a suite of state-of-the-art open-source and closed-source MLLMs using Gemini-2.5-Pro as the judge model for assessment \cite{comanici2025gemini}.
As shown in Table~\ref{tab:bechmark_mmmb_table}, InteractiveOmni-4B outperforms leading vision-language models such as Qwen2.5-VL-7B \cite{bai2025qwen25} and InternVL3-8B~\citep{zhu2025internvl3}, as well as  Omni-MLLMs including Qwen2.5-Omni-7B~\citep{xu2025qwen2} and GPT‑4o-mini~\citep{gpt4o}.
InteractiveOmni-8B further strengthens the multi-turn performance and is comparable to Gemini-2.5-Flash ~\citep{comanici2025gemini} (\textbf{58.17 vs. 60.84}), demonstrating the strong performance in long-term memory for historical image and text context.
To quantify the model performance degradation related to memory information, we conduct two types of evaluation: one measures accuracy based on the number of historical images to be recalled, and the other assesses accuracy according to the turn distance between critical historical turns and the final question.
As shown in Figure~\ref{fig:mmmb_performance_degradation}, the model's performance decreases as the turn distance increases. InteractiveOmni-4B maintains an accuracy of 40\% even with a turn distance of four. This demonstrates its robustness, which is comparable to the proprietary model such as Gemini-2.5-Flash, and significantly surpasses other open-source models like Qwen2.5-Omni-7B, InternVL3-8B, and Qwen2.5-VL-7B, which only achieve a score of around 20.
Additionally, all models exhibit a significant performance decline as the number of images to be memorized increases, highlighting a common weakness in the long-term memory of current MLLMs. For example, even Gemini-2.5-Flash achieves a score of only 20 under these conditions.


\begin{table*}[ht]
\centering
\caption{Performance evaluation of InteractiveOmni, proprietary and open-source models on the MMMB.}
\label{tab:bechmark_mmmb_table}
\resizebox{\textwidth}{!}{
\begin{tabular}{c|l|cccc}
\toprule
\textbf{Type}&\textbf{Model} & \textbf{Text Memory} & \textbf{Image Memory} & \textbf{Mixed Memory} & \textbf{Average}  \\
\midrule
\multirow{2}{*}{Proprietary}&GPT‑4o-mini~\citep{gpt4o} & 70.00 & 29.41 & 58.06 & 51.33\\
&Gemini-2.5-Flash~\citep{comanici2025gemini} & 75.76 & 40.19 & 70.97 & 60.84 \\
\midrule
\multirow{4}{*}{Open-source}
&InternVL3-8B~\citep{zhu2025internvl3} & 31.31& 15.69 & 33.87 & 25.86\\
&Qwen2.5-VL-7B~\citep{bai2025qwen25}   & 35.35 & 13.73 & 27.42 & 25.10 \\
&Qwen2.5-Omni-7B~\citep{xu2025qwen2}  &  32.32 & 9.80 & 40.32 & 25.48 \\
&InteractiveOmni-4B & \underline{70.71} & \underline{30.39} & \underline{59.68} & \underline{52.47}\\
&InteractiveOmni-8B & \textbf{72.73} & \textbf{40.20} & \textbf{64.52} & \textbf{58.17}\\
\bottomrule
\end{tabular}
}
\end{table*}




\begin{figure*}[htb]
    \centering
    \includegraphics[width=\linewidth]{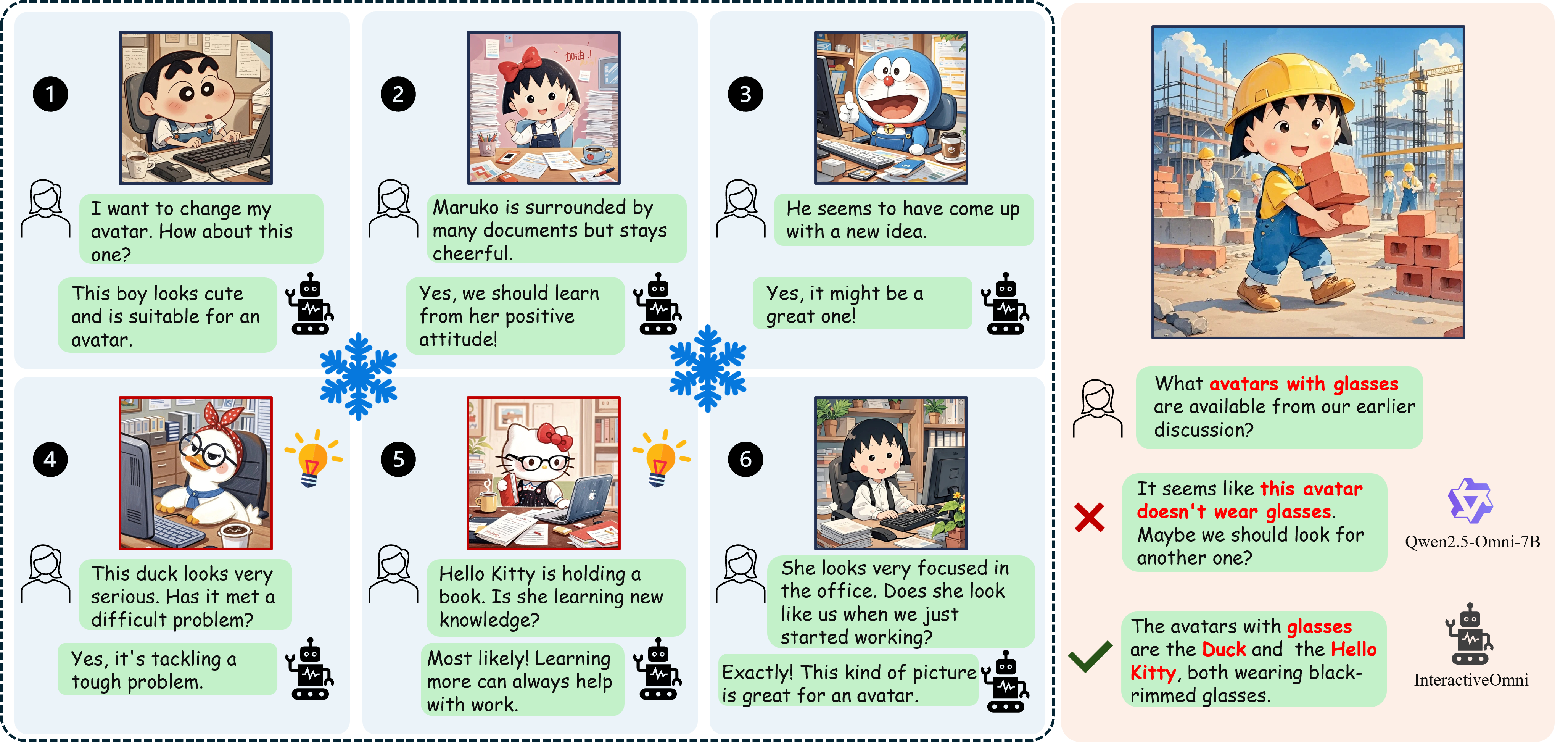}
    \caption{
    An example of multi-turn conversations requiring historical image context. InteractiveOmni demonstrates enhanced long-term memory performance for historical images compared to Qwen2.5-Omni-7B.
    }
    \label{fig:case_MMMB}
\end{figure*}

As shown in Figure~\ref{fig:case_MMMB}, we compare the performance of Qwen2.5-Omni-7B and InteractiveOmni in multimodal and multi-turn conversations. The results demonstrate that InteractiveOmni can accurately answer questions based on historical image information, showcasing its strong long-term memory capability.

\subsubsection{Multi-turn Speech Interaction Benchmark (MSIB)}
\label{subsec:msib}

\paragraph{Benchmark Introduction.} To comprehensively assess InteractiveOmni in realistic multi-turn speech dialogue scenarios, we propose the Multi-turn Speech Interaction Benchmark (MSIB). MSIB spans six measurable dimensions: \emph{basic conversational ability}, \emph{emotional expression capability}, \emph{speech rate control ability}, \emph{role-playing proficiency}, \emph{creative capacity}, and \emph{instruction-following ability}. 
This multi-faceted design enables a comprehensive evaluation of end-to-end audio dialogue systems.
The complete task formulations and evaluation protocol (including prompts, turn-structure, and model-as-judge rubric) are detailed in Appendix~\ref{sec:multi-turn_speech_benchmark}. 
We compare InteractiveOmni-4B and InteractiveOmni-8B against two leading audio-language models, Qwen2.5-Omni-7B~\cite{xu2025qwen2} and Kimi-Audio~\cite{kimiteam2025kimiaudiotechnicalreport}.

\paragraph{Automated Evaluation Results.} Table~\ref{tab:machine_eval} reports model-as-judge scores on the MSIB benchmark on a 1--5 scale. The automated evaluation results demonstrate the superior performance of InteractiveOmni across multiple dimensions of multi-turn speech interaction. 

\textbf{Content Quality Dominance.} InteractiveOmni-4B demonstrates clear superiority over both Qwen2.5-Omni-7B and Kimi-Audio in content quality, achieving the highest or second-highest scores across five out of six categories. It notably outperforms the baselines in Emotional Expression (3.97) and Role-Playing (3.80), exceeding the second-best baseline by substantial margins. The model also excels in Creative Capacity (3.83), highlighting its strong generative capabilities in empathetic and imaginative scenarios. The larger 8B variant further strengthens these results, leading in four of six content categories.

\textbf{Competitive Speech Quality.} In speech quality, InteractiveOmni-4B remains highly competitive, achieving the second-highest score in Emotional Expression (4.23) and outperforming Kimi-Audio across all categories. While Qwen2.5-Omni-7B leads in several speech tasks, the 4B model consistently surpasses Kimi-Audio, demonstrating a balanced and robust profile. The 8B model secures the top position in three categories including Basic Conversation (4.02) and Emotional Expression (4.26). 

InteractiveOmni-4B achieves an average score of 3.95, significantly outperforming both Qwen2.5-Omni-7B (3.58) and Kimi-Audio (3.65), underscoring its balanced and comprehensive capabilities across content and speech dimensions. The 8B variant further elevates this performance, attaining the highest overall score of 4.03 and leading in all average category scores, reflecting the scalability of the InteractiveOmni series.

\begin{table*}[t]
\centering
\caption{Evaluation of InteractiveOmni and baseline models on MSIB using model-as-judge. Best results are in \textbf{bold} and
second-best results are \underline{underlined}, and scores range from 1 to 5.}
\label{tab:machine_eval}
\setlength{\tabcolsep}{2pt}
\resizebox{\textwidth}{!}{
\begin{tabular}{ll|cccccccc}
\toprule
\textbf{Dimension} & \textbf{Model} & 
\makecell{Basic\\Conversation} & 
\makecell{Emotional\\Expression} & 
\makecell{Rate\\Control} & 
\makecell{Role\\Playing} & 
\makecell{Creative\\Capacity} & 
\makecell{Instruction\\Following} & 
\makecell{Overall} \\
\midrule
\multirow{3}{*}{Content Quality}
& Qwen2.5-Omni-7B \cite{xu2025qwen2} & 3.14 & 2.59 & 3.29 & 2.57 & 3.12 & 3.14 & 2.96  \\
& Kimi-Audio \cite{kimiteam2025kimiaudiotechnicalreport} & 3.38 & 2.98 & \textbf{4.10} & 2.83 & 3.44 & \underline{3.43} & 3.37  \\
& InteractiveOmni-4B & \underline{3.67} & \underline{3.97} & 3.90 & \underline{3.80} & \underline{3.83} & \textbf{3.81} & \underline{3.84}  \\
& InteractiveOmni-8B & \textbf{3.70} & \textbf{4.00} & \underline{3.92} & \textbf{4.03} & \textbf{4.05} & \underline{3.43} & \textbf{3.89}  \\
\midrule
\multirow{3}{*}{Speech Quality}
& Qwen2.5-Omni-7B \cite{xu2025qwen2} & \underline{3.98} & 4.13 & \textbf{4.41} & \textbf{4.33} & \textbf{4.22} & 4.00 & \textbf{4.19}  \\
& Kimi-Audio \cite{kimiteam2025kimiaudiotechnicalreport} & 3.64 & 4.03 & 3.92 & 3.90 & \underline{4.05} & \underline{4.10} & 3.93  \\
& InteractiveOmni-4B & 3.79 & \underline{4.23} & 4.16 & 3.93 & 4.02 & 4.00 & 4.05  \\
& InteractiveOmni-8B & \textbf{4.02} & \textbf{4.26} & \underline{4.22} & \underline{4.10} & \underline{4.05} & \textbf{4.33} & \underline{4.16}  \\
\midrule
\multirow{3}{*}{Average}
& Qwen2.5-Omni-7B  \cite{xu2025qwen2} & 3.56 & 3.36 & 3.85 & 3.45 & 3.67 & 3.57 & 3.58  \\
& Kimi-Audio \cite{kimiteam2025kimiaudiotechnicalreport} & 3.51 & 3.51 & 4.01 & 3.37 & 3.75 & 3.77 & 3.65  \\
& InteractiveOmni-4B & \underline{3.73} & \underline{4.10} & \underline{4.03} & \underline{3.87} & \underline{3.93} & \textbf{3.91} & \underline{3.95}  \\
& InteractiveOmni-8B & \textbf{3.86} & \textbf{4.13} & \textbf{4.07} & \textbf{4.07} & \textbf{4.05} & \underline{3.88} & \textbf{4.03}  \\
\bottomrule
\end{tabular}
}
\vspace{5pt}
\end{table*}

\begin{figure}[!h]
    \centering
    \begin{subfigure}{0.5\textwidth}
        \centering
        \includegraphics[width=\linewidth]{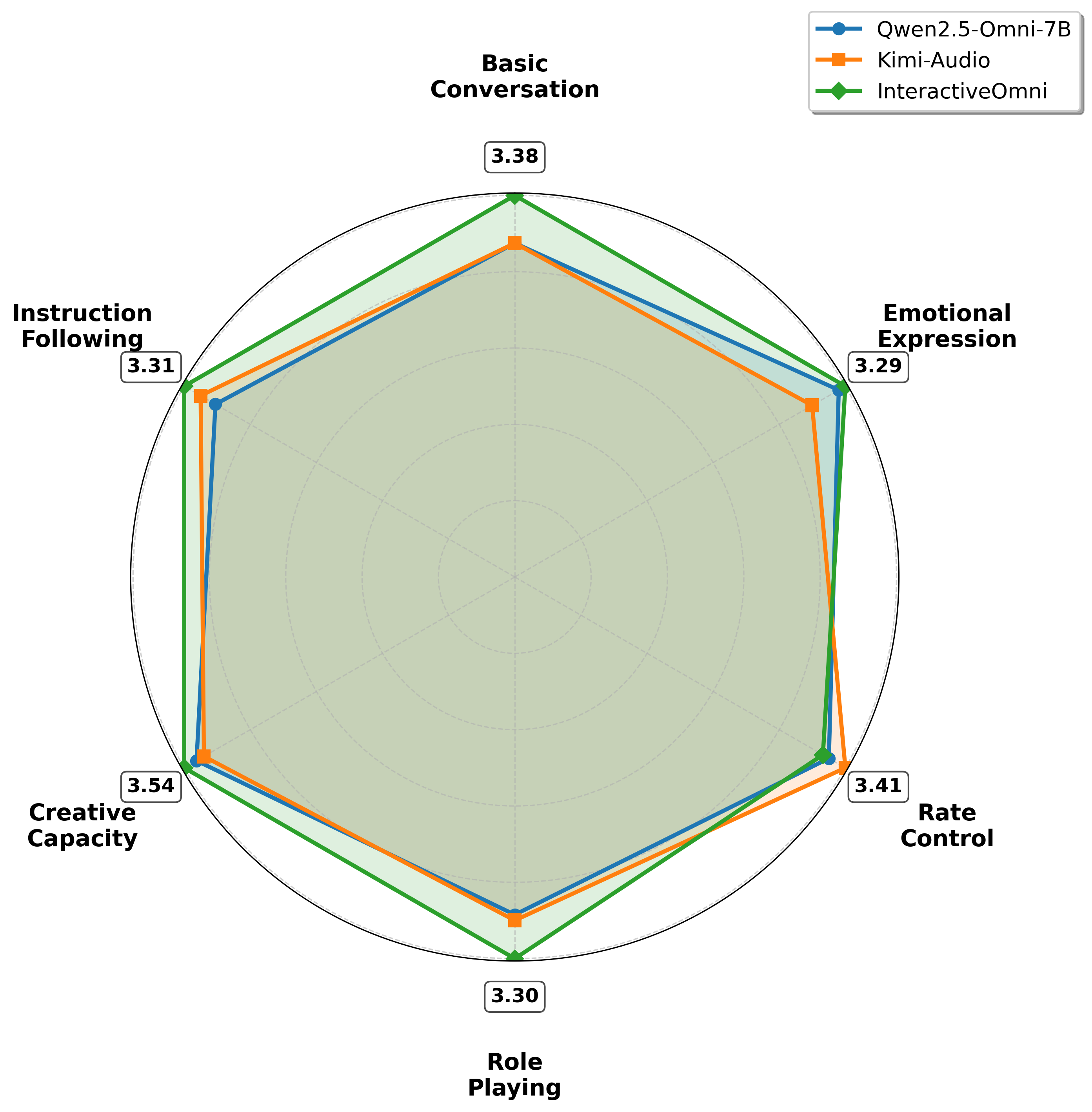}
        \label{fig:content_speech_average_quality_radar}
    \end{subfigure}
    \caption{Human evaluation of the speech-to-speech interactions on MSIB.}
    \label{fig:human_eval_radar}
\end{figure}


\paragraph{Human Evaluation.}  
As shown in Figure \ref{fig:human_eval_radar}, human evaluators rate the speech-to-speech conversations on a 1-5 Mean Opinion Score (MOS) scale. The results demonstrate that InteractiveOmni consistently outperforms the baselines across multiple dimensions of general conversational quality (simultaneously considering speech and content quality). Compared with Qwen2.5-Omni-7B and Kimi-Audio, InteractiveOmni achieves higher scores in \emph{Basic Conversation}, \emph{Emotional Expression}, \emph{Role-Playing}, \emph{Creative Capacity}, and \emph{Instruction Following}. These results confirm that InteractiveOmni delivers more expressive, coherent, and user-centric interactions, complementing automated metrics with clear human preference advantages.

\subsection{Open-source Benchmarks}

\subsubsection{Image Understanding Benchmarks}
To evaluate the comprehensive capabilities of the model in image understanding tasks, we conduct an extensive assessment on seven benchmarks: MMBench V1.1~\cite{mmbench}, MMStar~\cite{mmstar}, MMMU~\cite{mmmu}, MathVista~\cite{mathvista}, HallusionBench~\cite{hallusionbench}, AI2D~\cite{kembhavi2016diagram}, and OCRBench~\cite{liu2024ocrbench}. 
We compare the performance of our model with state-of-the-art vision-language models (VLMs) and omni-modal models of a similar scale, including InternVL3-8B~\cite{zhu2025internvl3}, InternVL3.5-8B~\cite{wang2025internvl35advancingopensourcemultimodal}, Qwen2.5-VL-7B~\cite{bai2025qwen25}, Qwen2.5-Omni-7B~\cite{xu2025qwen2} and GPT-4o-mini \cite{gpt4o}.
As shown in Table \ref{tab:image_understading}, InteractiveOmni demonstrates competitive performance with VLMs such as InternVL3-8B and Qwen2.5-VL-7B, and is superior to the open-source omni-modal model such as Qwen2.5-Omni-7B. Specifically, InteractiveOmni-8B outperforms all open-source models in HallusionBench, achieving a score of 61.3.
These results indicate that InteractiveOmni maintains robust image understanding capabilities and achieves leading performance in specific scenarios.


\begin{table*}[h]
\centering
\caption{Results on image understanding benchmarks. 
The score of other models is taken from the OpenCompass \cite{2023opencompass}. The best result is highlighted in \textbf{bold}, the second-best is \underline{underlined}.
}
\label{tab:image_understading}
\resizebox{\textwidth}{!}{
\setlength{\tabcolsep}{1.5pt}
\begin{tabular}{c|c|cccccccc}
\toprule
\textbf{Type}  & \textbf{Model}               &   \textbf{MMBench-V1.1} &   \textbf{MMStar} &   \textbf{MMMU} &  \textbf{MathVista} &   \textbf{HallusionBench} &   \textbf{AI2D} &   \textbf{OCRBench} &   \textbf{Avg}  \\
\midrule
\multirow{3}{*}{Visual} & InternVL3-8B~\cite{zhu2025internvl3}    & 82.1 & 68.7 & 62.2 & 70.5 & 49.0 & 85.1  & 88.4 &  72.3\\
                        & InternVL3.5-8B \cite{wang2025internvl35advancingopensourcemultimodal}  & 79.5 & 69.3 & 73.4 & 78.4 & 54.5 & 84.0  & 84.0 &  74.7\\
                        & Qwen2.5-VL-7B \cite{bai2025qwen25}   &  82.2 & 64.1 & 58.0 & 68.1 & 51.9 & 84.3 & 88.8  & 71.1 \\
\midrule
\multirow{6}{*}{Omni}  & GPT-4o-mini~\cite{gpt4o}     & 76.0 & 54.8 & \underline{60.0} & 52.5 & 46.1 & 77.8 & 78.5  & 63.7  \\
                        & VITA-1.5 \cite{fu2025vita}  & 76.8 & 60.2 & 52.6 & 66.2 & 44.6 & 79.2 & 74.1 & 64.8 \\
                        & Ming-Lite-Omni \cite{ai2025ming} & 80.8 & \underline{64.7} & 56.3 & \textbf{71.6} & \underline{55.0} & 83.1 & \textbf{88.4} & \underline{71.4} \\
                        & Qwen2.5-Omni-7B~\cite{xu2025qwen2} &  \underline{81.3} & 64.0 & 59.2 & 67.9 & 47.4 & \underline{83.2} & 83.4  & 69.5 \\
                        & InteractiveOmni-4B & 78.9 & 62.6 & 61.1 & 61.7 & 52.2 & 83.8 & 80.0 & 68.6 \\
                        & InteractiveOmni-8B      &  \textbf{81.4} & \textbf{66.8} & \textbf{66.9} &  \underline{68.0} & \textbf{61.3} & \textbf{84.3} & \underline{83.7}  & \textbf{73.2} \\
\bottomrule
\end{tabular}
}
\end{table*}

\begin{table*}[htb]
\centering
\caption{Results on video understanding benchmarks.}
\setlength{\tabcolsep}{3pt}
\resizebox{\textwidth}{!}{
\begin{tabular}{c|c|ccccc}
\toprule
\textbf{Type}  & \textbf{Model}  & 
\textbf{Video-MME(wo sub)} & 
\textbf{Video-MME(w sub)} & 
\textbf{MLVU(M-Avg)} & 
\textbf{LongVideoBench(val total)} & 
\textbf{Avg} \\
\midrule
\multirow{3}{*}{Visual} & InternVL3-8B~\cite{zhu2025internvl3} &  \textbf{66.3} & 68.9 & \underline{71.4} & 58.8 & 66.4  \\
&InternVL3.5-8B  \cite{wang2025internvl35advancingopensourcemultimodal} &  \underline{66.0} & 68.6 & 70.2 & \textbf{62.1} & \underline{66.7}  \\
&Qwen2.5-VL-7B~\cite{bai2025qwen25} &  65.1 & 71.6 & 70.2 & 56.0 & 64.5  \\
\midrule
\multirow{4}{*}{Omni} 
& GPT-4o-mini~\cite{gpt4o} &  64.8 & - & - & -   & - \\
&  Qwen2.5-Omni-7B~\cite{xu2025qwen2} &  64.3 & \textbf{72.4} & - & - & - \\
& InteractiveOmni-4B &  63.3 & 69.3 & 68.0 & 57.0 & 64.4  \\
& InteractiveOmni-8B &  \underline{66.0} & \underline{71.8} & \textbf{71.6} & \underline{59.1} & \textbf{67.1}  \\
\bottomrule
\end{tabular}
}
\label{tab:video_understanding}
\end{table*}

\begin{table*}[!h]
\centering
\small 
\caption{Results on ASR benchmarks. 
The best result is highlighted in \textbf{bold}, the second-best is \underline{underlined}, and results reproduced by ourselves are marked with *.
}
\begin{tblr}{
  column{1} = {c},
  cell{1}{3} = {c},
  cell{2}{1} = {r=8}{},
  cell{10}{1} = {r=6}{},
  cell{16}{1} = {r=6}{},
  cell{22}{1} = {r=6}{},
  cell{28}{1} = {r=7}{},
  cell{35}{1} = {r=4}{},
  hline{1,10,16,22,28,35,39} = {-}{0.08em},
  hline{2} = {-}{0.05em},
  hline{8,14,20,26,33,37} = {2-3}{0.03em},
}
\textbf{Datasets} & \textbf{Model}   & \textbf{Performance (WER)} $\downarrow$ \\

{\textbf{LibriSpeech~}~\cite{panayotov2015librispeech}\\ \textit{dev-clean} \textbar{} \textit{dev-other} \textbar{}\\
\textit{test-clean} \textbar{} \textit{test-other}} & Qwen2.5-Omni-7B~\cite{xu2025qwen2}  & 1.60 \textbar{} 3.50 \textbar{} 1.80 \textbar{} \underline{3.40}       \\
    & Qwen2-Audio~\cite{chu2024qwen2}      & \textbf{1.30} \textbar{} \underline{3.40} \textbar{} \underline{1.60} \textbar{} 3.60       \\
    & Step-Audio-Chat~\cite{huang2025stepaudiounifiedunderstandinggeneration}     & \;\;\;-\;\; \textbar{} \;\;\;-\;\; \textbar{} 3.19 \textbar{} 10.67         \\
    & Kimi-Audio~\cite{kimiteam2025kimiaudiotechnicalreport}    & \;\;\;-\;\; \textbar{} \;\;\;-\;\; \textbar{} \textbf{1.28} \textbar{} \textbf{2.42}         \\
    & Mini-Omni2~\cite{xie2024mini}       & 4.70 \textbar{} 9.40 \textbar{} 4.80 \textbar{} 9.80       \\
    & VITA-1.5~\cite{fu2025vita}         & 8.14 \textbar{} 18.41 \textbar{} 7.57 \textbar{} 16.57 \\
    & InteractiveOmni-4B        & 1.60 \textbar{} \textbf{3.38} \textbar{} 1.73 \textbar{} 3.69   \\
    & InteractiveOmni-8B        & \underline{1.45} \textbar{} \textbf{3.38} \textbar{} 1.64 \textbar{} 3.41   \\

{\textbf{WenetSpeech}~\cite{zhang2022wenetspeech}\\~\textit{test-net} \textbar{} \textit{test-meeting}}
    & Qwen2.5-Omni-7B~\cite{xu2025qwen2}  & 5.90 \textbar{} 7.70 \\
    & Qwen2-Audio*~\cite{chu2024qwen2}     & 10.60 \textbar{} 10.68 \\
    & Step-Audio-Chat~\cite{huang2025stepaudiounifiedunderstandinggeneration}     & 8.75 \textbar{} 9.52 \\
    & Kimi-Audio~\cite{kimiteam2025kimiaudiotechnicalreport}    & \underline{5.37} \textbar{} \underline{6.28} \\
    & InteractiveOmni-4B        & 5.40 \textbar{} 6.95 \\
    & InteractiveOmni-8B        & \textbf{5.04} \textbar{} \textbf{5.55} \\

\textbf{AISHELL-1}~\cite{bu2017aishell}
    & Qwen2.5-Omni-7B~\cite{xu2025qwen2}  & \underline{1.13} \\
    & Qwen2-Audio*~\cite{chu2024qwen2}     & 3.01 \\
    & Step-Audio-Chat~\cite{huang2025stepaudiounifiedunderstandinggeneration}     & 1.95 \\
    & Kimi-Audio~\cite{kimiteam2025kimiaudiotechnicalreport}    & \textbf{0.60} \\
    & InteractiveOmni-4B        & 1.21 \\
    & InteractiveOmni-8B        & 1.48 \\

\textbf{AISHELL-2 IOS}~\cite{du2018aishell2}
    & Qwen2.5-Omni-7B~\cite{xu2025qwen2}  & \underline{2.56} \\
    & Qwen2-Audio*~\cite{chu2024qwen2}     & 4.48 \\
    & Step-Audio-Chat~\cite{huang2025stepaudiounifiedunderstandinggeneration}     & 3.57 \\
    & Kimi-Audio~\cite{kimiteam2025kimiaudiotechnicalreport}    & \underline{2.56} \\
    & InteractiveOmni-4B        & 2.85 \\
    & InteractiveOmni-8B        & \textbf{2.18} \\

{\textbf{FLEURS}~\cite{conneau2023fleurs}\\~\textit{zh} \textbar{} \textit{en}}
    & Whisper-Large-V3~\cite{whisper} & 7.70 \textbar{} \textbf{4.10} \\
    & Qwen2.5-Omni-7B~\cite{xu2025qwen2}  & \underline{3.00} \textbar{} \textbf{4.10} \\
    & Qwen2-Audio*~\cite{chu2024qwen2}     & 7.50 \textbar{} 5.67 \\
    & Step-Audio-Chat~\cite{huang2025stepaudiounifiedunderstandinggeneration}     & 4.26 \textbar{} 8.56 \\
    & Kimi-Audio~\cite{kimiteam2025kimiaudiotechnicalreport}    & \textbf{2.56} \textbar{} 4.44 \\
    & InteractiveOmni-4B        & 3.86 \textbar{} 4.53 \\
    & InteractiveOmni-8B        & 3.49 \textbar{} \underline{4.14} \\

\textbf{ChildMandarin}~\cite{zhou2024childmandarin}
    & Qwen2.5-Omni-7B*~\cite{xu2025qwen2} & 19.34 \\
    & Qwen2-Audio*~\cite{chu2024qwen2}     & \underline{14.62} \\
    & InteractiveOmni-4B        & 17.21 \\
    & InteractiveOmni-8B        & \textbf{14.03}

\end{tblr}
\label{audio asr}
\end{table*}

\begin{table}[t]
\centering
\caption{Results on audio understanding tasks. The best result is highlighted in \textbf{bold}, the second-best is \underline{underlined}, and results reproduced by ourselves are marked with *.
}
\vspace{5pt}
\vspace{5pt}
\begin{tblr}{
  column{1} = {c},
  cell{2}{1} = {r=6}{},
  cell{8}{1} = {r=6}{},
  cell{14}{1} = {r=6}{},
  cell{20}{1} = {r=6}{},
  hline{1,8,14,20,26} = {-}{0.08em},
  hline{2} = {-}{0.05em},
  hline{6,12,18,24} = {2-3}{0.03em},
}
\textbf{Datasets}
& \textbf{Model}   & \textbf{Performance} $\uparrow$ \\

{\textbf{MMAU}~\cite{sakshi2024mmau}\\ \textit{Sound} \textbar{} \textit{Music} \textbar{} \\ \textit{Speech} \textbar{} \textit{Avg.}}
    & Qwen2.5-Omni-7B~\cite{xu2025qwen2}  & 67.78 \textbar{} 69.16 \textbar{} 59.76 \textbar{} 65.60 \\
    & Qwen2-Audio*~\cite{chu2024qwen2}     & 60.66 \textbar{} 58.08 \textbar{} 51.05 \textbar{} 56.6 \\
    & Kimi-Audio~\cite{kimiteam2025kimiaudiotechnicalreport}    & \textbf{73.27} \textbar{} 61.68 \textbar{} 60.66 \textbar{} 65.20 \\
    & MiDashengLM-7B~\cite{dinkel2025midashenglm}   & 68.47 \textbar{} 66.77 \textbar{} \underline{63.66} \textbar{} 66.30 \\
    & InteractiveOmni-4B        & \underline{70.87} \textbar{} \textbf{76.05} \textbar{} \textbf{69.07} \textbar{} \textbf{72.00} \\
    & InteractiveOmni-8B        & 69.07 \textbar{} \underline{73.05} \textbar{} 60.06 \textbar{} \underline{67.39} \\

{\textbf{AIR-Bench }~\cite{yang2024airbench}\\ \textit{Speech} \textbar{} \textit{Sound} \textbar{} \\ \textit{Music} \textbar{} \textit{Mixed-Audio} \\  \textbar{} \textit{Avg}}
    & Qwen2.5-Omni-7B*~\cite{xu2025qwen2} & \textbf{7.96} \textbar{} 6.82 \textbar{} 6.43 \textbar{} 6.69   \textbar{} 6.98  \\
    & Qwen2-Audio~\cite{chu2024qwen2}      & 7.18 \textbar{} 6.99 \textbar{} \textbf{6.79} \textbar{} \underline{6.77}  \textbar{} 6.93   \\
    & SALMONN~\cite{tang2023salmonn}      & 6.16 \textbar{} 6.28 \textbar{} 5.95 \textbar{} 6.08   \textbar{} 6.12  \\
    &   Phi-4-Multimodal~\cite{abouelenin2025phi}      & 7.47 \textbar{} 7.00 \textbar{} \underline{6.67} \textbar{} \textbf{6.78}  \textbar{} 6.98   \\
    & InteractiveOmni-4B        & \underline{7.82} \textbar{} \textbf{7.13} \textbar{} 5.91 \textbar{} 5.55   \textbar{} 6.60  \\
    & InteractiveOmni-8B        & 7.54 \textbar{} \underline{7.07} \textbar{} 6.00 \textbar{} 5.54  \textbar{} 6.54   \\

\textbf{MELD}~\cite{poria2018meld}
    & Qwen2.5-Omni-7B~\cite{xu2025qwen2}  & 57.00 \\
    & Qwen2-Audio~\cite{chu2024qwen2}      & 55.30 \\
    & Step-Audio-Chat~\cite{huang2025stepaudiounifiedunderstandinggeneration}      & 33.54 \\
    & Kimi-Audio~\cite{kimiteam2025kimiaudiotechnicalreport}    & \textbf{59.13} \\
    & InteractiveOmni-4B & 57.16 \\
    & InteractiveOmni-8B & \underline{57.55} \\

{\textbf{ClothoAQA}~\cite{lipping2022clothoaqa} \\ \textit{dev} \textbar{} \textit{test}}                                                    & Qwen2.5-Omni-7B~\cite{xu2025qwen2} & \underline{73.12} \textbar{} \underline{72.86} \\
    & Qwen2-Audio~\cite{chu2024qwen2}   & 72.63 \textbar{} 71.73 \\           
    & Step-Audio-Chat~\cite{huang2025stepaudiounifiedunderstandinggeneration}   & 44.98 \textbar{} 45.84 \\
    & Kimi-Audio~\cite{kimiteam2025kimiaudiotechnicalreport}   & \textbf{73.18} \textbar{} 71.24 \\
    & InteractiveOmni-4B & 71.91 \textbar{} 71.28 \\
    & InteractiveOmni-8B & 72.98 \textbar{} \textbf{74.49}

\end{tblr}
\label{audio comprehensive}
\end{table}

\subsubsection{Video Understanding Benchmarks}
To assess the video understanding capabilities, we conduct a thorough
evaluation on representative video understanding benchmarks including Video-MME~\cite{mmbench}, MLVU~\cite{mmstar}, and LongVideoBench~\cite{mmmu}. 
We compare InteractiveOmni with several state-of-the-art vision-language models, including InternVL3-8B \cite{zhu2025internvl3}, InternVL3.5-8B \cite{wang2025internvl35advancingopensourcemultimodal}, Qwen2.5-VL-7B \cite{bai2025qwen25}, as well as omni-modal models such as Qwen2.5-Omni-7B \cite{xu2025qwen2} and GPT-4o-mini~\cite{gpt4o}.
As shown in Table \ref{tab:video_understanding}, InteractiveOmni achieves competitive performance against existing vision-language models and outperforms Qwen2.5-Omni-7B across multiple benchmarks.
These results demonstrate that InteractiveOmni maintains robust and consistent performance across a diverse set of video understanding tasks.

\subsubsection{Audio Understanding Benchmarks}
To thoroughly assess the audio understanding capabilities of our model, we conduct extensive evaluations across a wide range of automatic speech recognition (ASR) and comprehensive audio understanding benchmarks. 
The ASR evaluations include LibriSpeech (dev-clean, dev-other, test-clean, test-other)~\cite{panayotov2015librispeech}, WenetSpeech (test-net, test-meeting)~\cite{zhang2022wenetspeech}, AISHELL-1 (test)~\cite{bu2017aishell}, AISHELL-2 iOS (test)~\cite{du2018aishell2}, FLEURS (zh, en)~\cite{conneau2023fleurs}, and ChildMandarin~\cite{zhou2024childmandarin},
and we use word error rate (WER) to evaluate the performance.
As shown in Table \ref{audio asr}, InteractiveOmni-4B achieves competitive performance with much larger specialized audio-language models such as Qwen2-Audio \cite{chu2024qwen2}, Step-Audio-Chat \cite{huang2025stepaudiounifiedunderstandinggeneration} and Kimi-Audio \cite{kimiteam2025kimiaudiotechnicalreport}.
Specifically, InteractiveOmni-8B surpasses all open-source omni-modal and audio-language models on the challenging WenetSpeech benchmark, attaining a score of \textbf{5.04} on test-net and \textbf{5.55} on test-meeting, demonstrating the superior performance of the audio understanding ability. In addition, InteractiveOmni achieves state-of-the-art performance on both the AISHELL-2 IOS and ChildMandarin benchmarks, demonstrating its strong capability in Mandarin comprehension.

Beyond speech recognition, we systematically evaluate the model across a wide range of audio domains, including environmental sound detection, music analysis, speech comprehension, emotion recognition, audio question answering, and vocal sound classification. 
These assessments are conducted using benchmark datasets such as  MMAU~\cite{sakshi2024mmau}, AIR-Bench~\cite{yang2024airbench}, MELD~\cite{poria2018meld}, and ClothoAQA~\cite{lipping2022clothoaqa}. 
As shown in Table~\ref{audio comprehensive}, the comprehensive evaluation framework provides a holistic characterization of the model’s audio perception capabilities, highlighting the proficiency of InteractiveOmni in capturing complex acoustic patterns, including the paralinguistic information and sound signals. 
Notably, InteractiveOmni-4B demonstrates exceptional parameter efficiency by surpassing all open-source 7B-sized models on the MMAU benchmark, achieving an average score of \textbf{72.00}.

\subsubsection{Omni-modal Understanding Benchmarks}
We evaluate the omni-modal benchmark OmniBench~\citep{li2024omnibench} to assess the omni-modal understanding capability of MLLMs, and compare InteractiveOmni with Qwen2.5-Omni-7B and other models. 
As shown in Table \ref{tab:omnibench}, InteractiveOmni-4B achieves state-of-the-art performance on OmniBench, attaining an average score of \textbf{59.19} that substantially exceeds other Omni models, demonstrating exceptional omni-modal understanding capabilities.

\begin{table}[h]
\centering
\caption{Performance of InteractiveOmni on OmniBench compared with leading MLLMs.}
\vspace{5pt}
\label{tab:omnibench}
\begin{tabular}{l|cccc}
\toprule
\textbf{Model} & \textbf{Speech} & \textbf{Sound Event} & \textbf{Music} & \textbf{Avg} \\
\midrule
Gemini-1.5-Pro~\citep{team2024gemini}  & 42.67 & 42.26 & 46.23 & 42.91 \\
MIO-Instruct~\citep{wang2024mio} (7B) & 36.96 & 33.58 & 11.32 & 33.80 \\
AnyGPT (7B)~\citep{zhan2024anygpt}  & 17.77 & 20.75 & 13.21 & 18.04 \\
video-SALMONN (13B)~\citep{sun2024video} &34.11 & 31.70 & \textbf{56.60} & 35.64 \\
UnifiedIO2-xlarge (3.2B)~\citep{lu2024unified} &39.56 & 36.98 & 29.25 & 38.00 \\
UnifiedIO2-xxlarge (6.8B)~\citep{lu2024unified} &34.24 & 36.98 & 24.53 & 33.98 \\
MiniCPM-o-2.6~\citep{minicpm-v} & - & - & - & 40.50 \\
Baichuan-Omni-1.5~\citep{li2025baichuan} & - & - & - &  42.90 \\
Qwen2.5-Omni-7B~\citep{xu2025qwen2} & 55.25 & 60.00 & 52.83 & 56.13 \\
\midrule
InteractiveOmni-4B & \textbf{60.70} & \underline{61.51} & 42.45 & \underline{59.19} \\
InteractiveOmni-8B & \underline{60.18} & \textbf{62.64} & \underline{55.66} & \textbf{60.33} \\
\bottomrule

\end{tabular}
\end{table}

\begin{table}[!h]
\centering
\caption{Results on speech-to-text question-answering benchmarks.}
\vspace{5pt}
\resizebox{\textwidth}{!}{
\begin{tabular}{clc}
\toprule
\textbf{Datasets} & \textbf{Model} & \textbf{Performance}
\\\midrule \multirow{10}{*}{\begin{tabular}[c]{@{}c@{}}\textbf{OpenAudioBench} \\ \textit{Reasoning QA} \textbar{} \textit{Llama Questions} \textbar{} \\ \textit{Web Questions} \textbar{} \textit{TriviaQA} \textbar{} \\ \textit{AlpacaEval} \textbar{} \textit{Avg} \end{tabular}}   
   & Qwen2-Audio~\citep{chu2024qwen2} & 42.77 \textbar{} 69.67 \textbar{} 45.20 \textbar{} 40.30 \textbar{} 57.19 \textbar{} 51.03
\\ & GLM-4-Voice~\citep{zeng2024glm4voiceintelligenthumanlikeendtoend} & 47.43 \textbar{} 76.00 \textbar{} 55.40 \textbar{} 51.80 \textbar{} 57.89 \textbar{} 57.70
\\ & VITA-1.5~\citep{fu2025vita} & 41.00 \textbar{} 74.20 \textbar{} 57.30 \textbar{} 46.80 \textbar{} 68.20 \textbar{} 57.50
\\ & Step-Audio-chat~\citep{huang2025stepaudiounifiedunderstandinggeneration} & 60.00 \textbar{} 72.33 \textbar{} \textbf{73.00} \textbar{} 56.80 \textbar{} 56.53 \textbar{} 63.73
\\ & Baichuan-Audio~\citep{li2025baichuanaudiounifiedframeworkendtoend} & 41.90 \textbar{} 78.40 \textbar{} 64.50 \textbar{} 61.70 \textbar{} \underline{77.40} \textbar{} 64.78
\\ & Kimi-Audio~\citep{kimiteam2025kimiaudiotechnicalreport} & 58.02 \textbar{} \underline{79.33} \textbar{} 70.20 \textbar{} \underline{62.10} \textbar{} 75.73 \textbar{} 69.08
\\ & MiniCPM-o-2.6~\citep{minicpm-v} & 38.60 \textbar{} 77.80 \textbar{} 68.60 \textbar{} 61.90 \textbar{} 51.80 \textbar{} 59.74
\\ & Baichuan-Omni-1.5~\citep{li2025baichuan} & 50.00 \textbar{} 78.50 \textbar{} 59.10 \textbar{} 57.20 \textbar{} \textbf{77.90} \textbar{} 64.54
\\ & Qwen2.5-Omni-7B~\citep{xu2025qwen2} & 63.76 \textbar{} 75.33 \textbar{} 62.80 \textbar{} 57.06 \textbar{} 72.76 \textbar{} 66.34
\\ & InteractiveOmni-4B & \underline{69.11} \textbar{} \underline{79.33} \textbar{} 65.80 \textbar{} 56.40 \textbar{} 74.87 \textbar{} \underline{69.10}
\\ & InteractiveOmni-8B & \textbf{71.68} \textbar{} \textbf{80.67} \textbar{} \underline{70.30} \textbar{} \textbf{66.50} \textbar{} 74.57 \textbar{} \textbf{72.74}
\\\midrule \multirow{10}{*}{\begin{tabular}[c]{@{}c@{}}\textbf{VoiceBench} \\ \textit{AlpacaEval} \textbar{} \textit{CommonEval} \textbar{} \\ \textit{WildVoice} \textbar{} \textit{SD-QA} \textbar{} \textit{MMSU} \end{tabular}}   
   & Qwen2-Audio~\citep{chu2024qwen2} & 3.69 \textbar{} 3.40 \textbar{} 3.01 \textbar{} 35.35 \textbar{} 35.43
\\ & GLM-4-Voice~\citep{zeng2024glm4voiceintelligenthumanlikeendtoend} & 4.06 \textbar{} 3.48 \textbar{} 3.18 \textbar{} 43.31 \textbar{} 40.11
\\ & VITA-1.5~\citep{fu2025vita} & 4.21 \textbar{} 3.66 \textbar{} 3.48 \textbar{} 38.88 \textbar{} 52.15
\\ & Step-Audio-chat~\citep{huang2025stepaudiounifiedunderstandinggeneration} & 3.99 \textbar{} 2.99 \textbar{} 2.93 \textbar{} 46.84 \textbar{} 28.72
\\ & Baichuan-Audio~\citep{li2025baichuanaudiounifiedframeworkendtoend} & 4.41 \textbar{} 4.08 \textbar{} 3.92 \textbar{} 45.84 \textbar{} 53.19
\\ & Kimi-Audio~\citep{kimiteam2025kimiaudiotechnicalreport} & 4.46 \textbar{} 3.97 \textbar{} 4.20 \textbar{} \textbf{63.12} \textbar{} 62.17
\\ & MiniCPM-o-2.6~\citep{minicpm-v} & 4.42 \textbar{} 4.15 \textbar{} 3.94 \textbar{} 50.72 \textbar{} 54.78
\\ & Baichuan-Omni-1.5~\citep{li2025baichuan} & \underline{4.50} \textbar{} 4.05 \textbar{} \underline{4.06} \textbar{} 43.40 \textbar{} 57.25
\\ & Qwen2.5-Omni-7B~\citep{xu2025qwen2} & 4.50 \textbar{} 3.84 \textbar{} 3.89 \textbar{} \underline{56.40} \textbar{} 61.32
\\ & InteractiveOmni-4B & 4.27 \textbar{} \underline{4.20} \textbar{} 3.94 \textbar{} 41.41 \textbar{} \underline{63.24}
\\ & InteractiveOmni-8B & \textbf{4.61} \textbar{} \textbf{4.34} \textbar{} \textbf{4.21} \textbar{} 44.67 \textbar{} \textbf{65.26}
\\ \midrule \multirow{10}{*}{\begin{tabular}[c]{@{}c@{}}\textbf{VoiceBench} \\ \textit{OpenBookQA} \textbar{} \textit{IFEval} \textbar{} \\ \textit{BBH} \textbar{} \textit{AdvBench} \textbar{} \textit{Avg}  \end{tabular}}   
   & Qwen2-Audio~\citep{chu2024qwen2} & 49.01 \textbar{} 54.70 \textbar{} 22.57 \textbar{} 98.85	\textbar{} 55.32
\\ & GLM-4-Voice~\citep{zeng2024glm4voiceintelligenthumanlikeendtoend} & 52.97 \textbar{} 52.80 \textbar{} 24.91 \textbar{} 88.08	\textbar{} 57.40
\\ & VITA-1.5~\citep{fu2025vita} & 71.65 \textbar{} 55.30 \textbar{} 38.14 \textbar{} 97.69	\textbar{} 64.53
\\ & Step-Audio-chat~\citep{huang2025stepaudiounifiedunderstandinggeneration} & 31.87 \textbar{} 50.60 \textbar{} 29.19 \textbar{} 65.77	\textbar{} 50.13
\\ & Baichuan-Audio~\citep{li2025baichuanaudiounifiedframeworkendtoend} & 71.65 \textbar{} 54.80 \textbar{} 50.31 \textbar{} 99.42	\textbar{} 69.27
\\ & Kimi-Audio~\citep{kimiteam2025kimiaudiotechnicalreport} & \underline{83.52} \textbar{} \underline{69.70} \textbar{} \textbf{61.10} \textbar{} \textbf{100.0}	\textbar{} \textbf{76.91}
\\ & MiniCPM-o-2.6~\citep{minicpm-v} & 78.02 \textbar{} 60.40 \textbar{} 49.25 \textbar{} 97.69	\textbar{} 71.23
\\ & Baichuan-Omni-1.5~\citep{li2025baichuan} & 74.51 \textbar{} 62.70 \textbar{} 54.54 \textbar{} 97.31	\textbar{} 71.32
\\ & Qwen2.5-Omni-7B~\citep{xu2025qwen2} & 80.90 \textbar{} 66.70 \textbar{} 53.50 \textbar{} 99.20	\textbar{} 73.60
\\ & InteractiveOmni-4B & 82.64 \textbar{} 55.90 \textbar{} \underline{60.90} \textbar{} \underline{99.62} \textbar{} 73.10
\\ & InteractiveOmni-8B & \textbf{86.37} \textbar{} \textbf{73.30} \textbar{} 57.99 \textbar{} 99.42 \textbar{} \underline{76.69}
\\
\bottomrule

\end{tabular}}
\label{tab:speechqa}
\end{table}

\subsubsection{Speech-to-text Benchmarks}
To assess the speech understanding and speech-based question-answering capabilities, we evaluate InteractiveOmni on the following benchmarks:  OpenAudioBench \citep{li2025baichuanaudiounifiedframeworkendtoend} and VoiceBench \citep{chen2024voicebench}. 
 As shown in Table \ref{tab:speechqa}, our model performs excellently on the speech-to-text question-answering benchmarks, outperforming recent open-source audio language models and omni models. InteractiveOmni-4B achieves an average score of \textbf{69.10} on OpenAudioBench, significantly outperforming Kimi-Audio \cite{kimiteam2025kimiaudiotechnicalreport}, Step-Audio-chat \cite{huang2025stepaudiounifiedunderstandinggeneration} and Qwen2.5-Omni-7B \cite{{xu2025qwen2}}. These voice-chat benchmarks reflect our model’s substantial progress in diversified speech interaction.

\subsubsection{Text-to-Speech Benchmarks}
To evaluate the speech generation capability of InteractiveOmni, we conducte a comparative study on the Seed-TTS test set \cite{anastassiou2024seed} against both a state-of-the-art TTS system and the omni-modal models. The Seed-TTS benchmark includes Seed test-zh, test-en, and test-hard, covering diverse input texts and reference speeches across multiple domains. As shown in Table~\ref{tab:tts_benchmark}, compared with the omni-modal models such as Ming-Lite-Omni \cite{ai2025ming} and Qwen2.5-omni-7B$_{\text{icl}}$ \cite{xu2025qwen2}, InteractiveOmni-4B achieves considerably better performance on Seed-TTS-test-zh, reaching a level comparable to highly professional TTS systems.


\begin{table}[h]
\centering
\caption{Performance on the Seed-TTS benchmark, as measured by WER$\downarrow$.}
\small
\label{tab:tts_benchmark} 
\begin{tabular}{l|l|ccc}
\toprule
\textbf{Type} & \textbf{Model} & \textbf{test-zh} & \textbf{test-en}  & \textbf{test-zh-hard} \\
\midrule
\multirow{4}{*}{TTS} &MaskGCT \cite{wang2024maskgct} & 2.27 & 2.62 & 10.27 \\
&SeedTTS \cite{anastassiou2024seed} & 1.12 & 2.25 & 7.59 \\
&CosyVoice 2 \cite{du2024cosyvoice2scalablestreaming} & 1.45 & 2.57 & 6.83  \\
\midrule
\multirow{5}{*}{MLLM} &MinMo~\cite{chen2025minmomultimodallargelanguage} & 2.48 & 2.90 & - \\
&Ming-Lite-Omni \cite{ai2025ming} & 1.69 & 4.31 & - \\
&Qwen2.5-Omni-7B \cite{xu2025qwen2} & 1.70 & 2.72 & {7.97} \\
& InteractiveOmni-4B  & \textbf{1.37} & 3.73 & 8.02 \\
& InteractiveOmni-8B  & 1.56 & \textbf{2.33} & \textbf{7.92} \\
\bottomrule
\end{tabular}
\end{table}

To further assess how the models handle nuanced and semantically complex texts, we conduct evaluations on EmergentTTS-Eval\cite{manku2025emergenttts}, a comprehensive benchmark covering six challenging TTS scenarios: emotions, paralinguistics, foreign words, syntactic complexity, complex pronunciation (e.g., URLs, formulas), and questions. 
As shown in Table~\ref{tab:tts_emergent}, InteractiveOmni-4B and InteractiveOmni-8B achieve an overall WER of 22.04 and 18.07, respectively, surpassing all other models.
Furthermore, it reaches state-of-the-art performance in several sub-categories, including emotions, questions, paralinguistics, and complex pronunciation, surpassing leading omni-modal models.

\begin{table*}[h]
\centering
\caption{Performance on the EmergentTTS benchmark, as measured by WER$\downarrow$. The results for competing models are drawn from the EmergentTTS-Eval \cite{manku2025emergenttts}.}
\label{tab:tts_emergent}
\setlength{\tabcolsep}{3pt}
\resizebox{\textwidth}{!}{
\begin{tabular}{l|cccccccc}
\toprule
\textbf{Model} & 
\makecell{\textbf{Overall}} &
\makecell{\textbf{Emotions}} & 
\makecell{\textbf{Foreign}\\\textbf{Words}} & 
\makecell{\textbf{Paralinguistics}} & 
\makecell{\textbf{Complex}\\\textbf{Pronunciation}} & 
\makecell{\textbf{Questions}} & 
\makecell{\textbf{Syntactic}\\\textbf{Complexity}} \\
\midrule
VITS-VCTK~\cite{kim2021conditional} &  27.45 & 16.34 & 47.45 & 51.12 & \textbf{44.30} & 17.82 & 2.37  \\
Tortoise-TTS~\cite{betker2023betterspeechsynthesisscaling} &  28.62 & 13.04 & 29.61 & 64.93 & 51.87 & 10.44 & 6.35  \\
Sesame1B~\cite{csm2025} & 32.32 & 17.07 & 45.27 & 49.63 & 80.97 & 2.74 & 4.30  \\
MiniCPM-o-2.6~\cite{minicpm-v} &  31.40 & 12.36 & 33.46 & 58.48 & 82.15 & 5.21 & 3.08  \\
Qwen2.5-Omni-7B~\cite{xu2025qwen2} &  26.58 & 1.22 & \textbf{26.98} & 57.48 & 64.07 & 12.77 & 1.66  \\
\midrule
InteractiveOmni-4B &  22.04 & 1.59 & 29.75 & 33.37 & 66.36 & 1.67 & 5.19  \\
InteractiveOmni-8B &  \textbf{18.07} & \textbf{1.05} & 28.34 & \textbf{26.68} & {54.37} & \textbf{1.09} & \textbf{1.44}  \\
\bottomrule
\end{tabular}
}
\end{table*}

\subsubsection{Spoken Dialogue Benchmarks}
We evaluate the end-to-end spoken dialogue capabilities of InteractiveOmni based on the benchmarks: Llama Question \cite{nachmani2023spoken}, Web Question \cite{berant2013semantic}, TriviaQA \cite{joshi2017triviaqa}, and AlpacaEval \cite{li2023alpacaeval}. A comparison of the speech interaction abilities of speech LLMs and omni models is shown in Table~\ref{tab:speechqa_s2s}. InteractiveOmni achieves almost state-of-the-art performance across all four benchmarks on the speech-to-text and speech-to-speech evaluations, indicating its strong capabilities in handling a wide range of conversational scenarios.
These results show that InteractiveOmni excels in speech understanding and stable speech generation on end-to-end speech interaction scenarios.

\begin{table*}[t]
\centering
\caption{The performance of the speech LLMs on spoken dialogue benchmarks. S2T and S2S represent the speech-to-text and the speech-to-speech performance, respectively. Results reproduced by ourselves are marked with *.}
\label{tab:speechqa_s2s}
\resizebox{0.8\textwidth}{!}{
\begin{tabular}{l|cccccccc} 
    \toprule
    \multirow{2}{*}{Model} & \multicolumn{2}{c}{LlamaQuestion} & \multicolumn{2}{c}{WebQuestion} & \multicolumn{2}{c}{TriviaQA} & \multicolumn{2}{c}{AlpacaEval} \\
     & S2T$\uparrow$ & S2S$\uparrow$ & S2T$\uparrow$ & S2S$\uparrow$& S2T$\uparrow$ & S2S$\uparrow$ & S2T$\uparrow$ & S2S$\uparrow$ \\
    \midrule
    Moshi~\cite{defossez2024moshi} & 62.3 & 21.0 & 26.6 & 9.2 & 22.8 &  7.3 & - & - \\
    GLM-4-Voice~\cite{zeng2024glm4voiceintelligenthumanlikeendtoend} & 64.7 &  50.7 & 32.2 &  15.9 &  39.1 & 26.5 & - & - \\
    Kimi-Audio*~\cite{kimiteam2025kimiaudiotechnicalreport} & \textbf{82.0} & 66.6 & \underline{69.0} & 60.6 & \underline{64.2} & 52.8 & 58.2 & 36.4 \\
    Freeze-Omni~\cite{wang2024freeze} & 72.0 & - & 44.7 & - & 53.8 & - & - & - \\
    VITA-1.5~\cite{fu2025vita} & 74.2 & - & 57.3 & - & 46.8 & - & \underline{68.2} & - \\
    MiniCPM-o-2.6~\cite{minicpm-v} & 77.8 & - & 68.6 & - & 61.9 & - & 51.8 & - \\
    LLaMA-Omni2-7B~\cite{fang2025llamaomni2llmbasedrealtimespoken} & 70.3 & 60.7 & 34.5 & 31.3 & - & - & - & - \\
   Qwen2.5-Omni-7B*~\cite{xu2025qwen2} & 77.6 & \textbf{74.6} & 65.8 & \textbf{64.7} & 57.4 & \underline{55.9} & 56.1 & \underline{49.6} \\
    \midrule
    InteractiveOmni-4B & 76.3 & 65.3 & 64 & 58.7 & 53.1 & 43.8 & 53.8 & 46.4 \\
    InteractiveOmni-8B & \underline{81.0} & \underline{69.0} & \textbf{71.3} & \underline{64.6} & \textbf{66.8} & \textbf{56.3} & \textbf{74.3} & \textbf{58.1} \\
    \bottomrule
\end{tabular}
}
\end{table*}

\section{Related works}
\label{sec:related_works}

\subsection{Vision-language Models}
VLMs extend traditional LLMs by integrating vision and language understanding within a unified framework. By jointly processing textual and visual input, VLMs enable complex cross-modal reasoning tasks such as image captioning, visual question answering, and multimodal dialogue. 
They typically leverage pre-trained vision encoders, like CLIP~\cite{clip} or ViT~\cite{vit}, combined with powerful language backbones to align heterogeneous representations in a shared semantic space. 
Early multimodal models such as Flamingo~\cite{flamingo} and BLIP-2~\cite{blip2} focused on bridging vision encoders with language models through pre-training and efficient alignment strategies, enabling tasks such as captioning and visual question answering (VQA). Following this line of work, a series of instruction-tuned vision-language models emerged almost simultaneously, including Instruct-BLIP~\cite{instructblip}, LLaVA~\cite{llava}, MiniGPT-4~\cite{minigpt-4}, and mPLUG-Owl~\cite{mplug-owl}. These models adopt a common paradigm of leveraging large language models combined with vision encoders, and are trained on instruction-following datasets to enable effective multimodal alignment. Closed-source general-purpose models, such as GPT-4V~\cite{gpt4v} and Google’s Gemini~\cite{gemini}, extend multimodal capabilities beyond research prototypes by integrating text, vision, and other modalities within unified frameworks. 
Recent work like  Qwen2.5-VL~\cite{bai2025qwen25}, InternVL3.5~\cite{wang2025internvl35advancingopensourcemultimodal}, Seed1.5-VL~\cite{guo2025seed15vltechnicalreport}, GLM-4.1V~\cite{hong2025glm}, Kimi-VL~\cite{kimiteam2025kimivltechnicalreport} and MiMo-VL~\cite{coreteam2025mimovltechnicalreport}
primarily focused on native resolution, Mixture-of-Experts architecture, visual reasoning capabilities and reinforcement learning. In addition, to fully unleash the potential of the model, they collected a large amount of high-quality data, including available open-source datasets as well as carefully designed and curated in-house data. These optimizations and improvements have led to significant performance gains across a wide range of vision-related tasks, including OCR, general question answering, and visual reasoning.

\subsection{ Speech-to-Speech Dialogue Models}
Against the backdrop of the rapid evolution of LLM, both academia and industry have placed high expectations on the development of speech-to-speech models. Traditionally, end-to-end speech systems are constructed by sequentially integrating an automatic speech recognition module, an LLM, and a text-to-speech module, which is constrained by the high latency, insufficient paralinguistic perception, and error propagation across cascaded modules.
To address these challenges, several new paradigms have been proposed. For example, Twist \cite{hassid2024textuallypretrainedspeechlanguage} introduces a framework that enables pretrained textual language models to directly generate speech, thereby bridging the gap between text-based reasoning and spoken interaction. 
Moshi \cite{defossez2024moshi} proposes a full-duplex end-to-end spoken dialogue system that employs a multi-stream output mechanism to simultaneously produce audio and text tokens.

 Several speech-to-speech models have explored training strategies with curated training data to enhance the performance of end-to-end spoken dialogue systems. 
 GLM-4-Voice \cite{zeng2024glm4voiceintelligenthumanlikeendtoend} leverages interleaved data during pre-training to support text-guided interleaved speech generation. 
 MinMo \cite{chen2025minmomultimodallargelanguage} demonstrates the feasibility of end-to-end language systems by employing a multi-stage training strategy on 1.4 million hours of data. 
 Baichuan-Audio \cite{li2025baichuanaudiounifiedframeworkendtoend} adopts a multi-codebook discretization method to preserve both semantic and acoustic information, thereby enabling effective modeling of speech within the LLM. 
 Furthermore, LLaMA-Omni2 \cite{fang2025llamaomni2llmbasedrealtimespoken} explores techniques for integrating discrete text embeddings with continuous hidden representations.
Recently,  Kimi-Audio \cite{kimiteam2025kimiaudiotechnicalreport} achieves state-of-the-art performance across multiple speech and audio understanding benchmarks. 
Step-Audio 2 \cite{wu2025stepaudio2technicalreport} further advances the emotional and paralinguistic expressivity of end-to-end spoken dialogue systems by incorporating chain-of-thought reasoning and reinforcement learning. These studies achieve more natural speech-to-speech human-machine communication with end-to-end training.

\subsection{Omni-modal Large Language Model} 
To achieve human-like omni-modal interactive experience, the MLLMs have propelled the development of the Omni-modal MLLM, which can process omni-modality including image, video, audio, and text, such as GPT-4o \cite{gpt4o} and Gemini \cite{comanici2025gemini}.  
Compared with VLMs and ALMs, the Omni-MLLM integrates data from more modalities, enabling it to learn richer contextual information and gain a deeper understanding of the latent relationships between different modalities. 
VITA \cite{fu2024vita} achieves the omni-modal understanding capability, which can simultaneously process the video, image, text, and audio modalities towards the natural human-computer experience.
Mini-Omni2 \cite{xie2024miniomni2opensourcegpt4ovision} proposes the multi-modal model as a visual voice assistant to achieve the audio-visual interaction similar to the functionality of GPT-4o.
Ming-Omni \cite{ai2025ming} proposes a unified architecture capable of processing images, audio, video and text, while generating speech and images.
Qwen2.5-Omni \cite{xu2025qwen2} introduces an end-to-end model that can perceive all modalities and generate text and speech in a streaming fashion.

For the unified understanding and generation, the speech modality can be represented by the discrete audio token or continuous audio feature. 
Models based on discretized audio encoding \cite{defossez2024moshi,zeng2024glm4voiceintelligenthumanlikeendtoend,zhang2025omniflattenendtoendgptmodel} expand tokens into the vocabulary of the LLM to achieve unified understanding and generation of omni-modalities, while the training of the model usually requires a large amount of cross-modal data. 
In contrast, models based on continuous audio feature encoding \cite{fang2025llamaomni2llmbasedrealtimespoken,fang2025llamaomniseamlessspeechinteraction,wang2024freeze,xie2024miniomnilanguagemodelshear,chen2025minmomultimodallargelanguage} can maximize the preservation of the basic capabilities of the LLM. 
Thus, the omni-modal alignment and the design of a unified understanding and generation architecture still present several challenges. In addition, these omni-modal MLLMs show poor performance in multi-turn interaction, failing to achieve a natural, human-like conversational flow.

\section{Conclusions}
\label{sec:conlusions}
In this work, we present InteractiveOmni, a unified, open-source omni-modal large language model that seamlessly integrates comprehensive multi-modal understanding with natural speech generation, demonstrating superior performance on multi-turn interaction tasks. Our unified framework successfully integrates the processing of text, image, audio, and video inputs, and directly generates coherent text and speech, enabling a truly seamless and intelligent interactive experience.

Combining omni-modal pre-training for foundational modality alignment and post-training for omni-modal understanding and audio-visual interaction, we effectively address the critical challenge of cross-modal synergy. 
Furthermore, the meticulous curation of high-quality, multi-turn dialogue data is essential in endowing InteractiveOmni with robust long-term memory and contextual awareness, enabling interactions that are significantly more natural and intelligent.
InteractiveOmni demonstrates a clear superiority over comparable models in our newly constructed multi-turn benchmarks, showcasing its advanced capabilities in maintaining context and memory in complex dialogues. 
Moreover, it achieves state-of-the-art performance against similarly sized MLLMs across a suite of 
mainstream open-source benchmarks for image, audio, and video understanding, as well as speech conversation, proving its robustness and versatility.

InteractiveOmni lays a strong foundation for the next generation of multi-modal AI assistants. Our future work includes enhancing the model's efficiency for real-time interaction and expanding its capacity to comprehend more complex, abstract inter-modal relationships, paving the way for a more authentic and human-like user experience.

\bibliographystyle{plain}
\bibliography{neurips_2025}

\clearpage
\newpage
\appendix

\section{Evaluation Details of Multi-turn Speech Interaction Benchmark}
\label{sec:multi-turn_speech_benchmark}

\subsection{Data Preparation and Inference Pipeline}

We construct the multi-turn speech interaction benchmark to assess the model's core capability of speech-to-speech multi-turn interaction. 
We construct a total of 244 multi-turn dialogues with each dialogue consisting of 2 to 10 turns, covering six dimensions, such as basic conversational ability, emotional expression capability, speech rate control ability, role-playing proficiency, creative capacity, and instruction-following ability. The data construction and evaluation process consists of the following three steps: 

\begin{enumerate}
    \item \textbf{Text-based Dialogue Construction:} For each dimension, we first use an LLM to generate multi-turn spoken dialogues, which are then manually revised by annotators. Only the final turn of each dialogue was evaluated, while all previous turns were treated as pre-defined conversational history.
    \item \textbf{Speech-based Dialogue Construction:} For text-based multi-turn dialogues, we employ a high-quality TTS system to convert the textual prompts into speech, thereby generating speech-based multi-turn conversations to evaluate the model's speech interaction capabilities.
    \item \textbf{Inference:} When evaluating the model's performance, only the final dialogue turn is assessed, while all preceding turns are treated as historical context.
\end{enumerate}

\subsection{Human Evaluation}
To assess the model's performance in end-to-end speech interaction, expert evaluators rate the generated speech on a 1-5 Mean Opinion Score (MOS) scale. The evaluation covers both speech content and speech quality based on the detailed scoring rubric provided below.

\begin{mdframed}[
    backgroundcolor=green!10,
    linecolor=green!50!black,
    linewidth=2pt,
    roundcorner=10pt
]
\textbf{Speech Quality}

\qquad 1: Unintelligible or extremely difficult to understand. Critical flaws: Extremely loud background noise severely impacts comprehension; completely robotic/electronic voice with utterly stiff intonation.
    
\qquad 2: Sounds like a robot reading a script. Audible but very monotonous voice with zero emotional expression, similar to navigation systems or early AI voices.
    
\qquad 3: Sounds human but lacks emotion or has noticeable flaws. Clear speech that sounds human, but has flat intonation without emotional variation. May contain noticeable defects (e.g., occasional weird pronunciation, slightly muffled sound).
    
\qquad 4: Basically indistinguishable from human but not perfect. Clear speech resembling human speaking with basic intonation variations and some emotion. May have very minor flaws that don't affect the overall listening experience.
    
\qquad 5: Perfect and indistinguishable from human. Completely clear voice that not only sounds human but is emotionally rich with proper prosody and modulation, showing full expressiveness.

\textbf{Content Quality}

\qquad \textbf{Step 1: Determine Category}

\qquad \qquad - If BOTH content and attributes are poor → Score 1 (End evaluation)
        
\qquad \qquad - If ONE aspect is good but the other is poor → Proceed to 2-3 score range
        
\qquad \qquad - If BOTH content and attributes are good → Proceed to 4-5 score range

\qquad \textbf{Step 2: Detailed Scoring}

\qquad \qquad For 2-3 Range (One aspect deficient):
        
\qquad \qquad \qquad - Score 2: If attributes are severely mismatched (even if content is acceptable). Examples: Required role play completely missed; requested slow speed but too fast to understand; emotional tone completely wrong.
        
\qquad \qquad \qquad - Score 3: If attributes are relatively well-matched (regardless of content quality). Examples: Emotional expression mostly appropriate; role play generally convincing despite content issues.

\qquad \qquad For 4-5 Range (Both aspects good):
        
\qquad \qquad \qquad - Score 4: Accurately solves the problem and clearly meets attribute requirements. Standard, satisfactory completion.
        
\qquad \qquad \qquad - Score 5: Perfectly solves the problem (may include extra value) and demonstrates highly precise attribute fulfillment. Examples: Role play is vivid and authentic; instruction following is exceptionally well-executed.
\end{mdframed}

\subsection{Automated Machine Evaluation.}
To enable scalable assessment of multi-turn speech interactions, we implement an automated scoring pipeline leveraging LLM as a judge. We employ \textbf{Gemini-2.5-Pro} as the judge model, owing to its state-of-the-art multi-modal understanding capabilities and demonstrated proficiency in complex reasoning tasks. The prompt for the judge model is given as follows:

\begin{mdframed}[
    backgroundcolor=green!10,
    linecolor=green!50!black,
    linewidth=2pt,
    roundcorner=10pt
]

You are an AI audio assistant acting as a strong reward model for evaluating an end-to-end (speech-to-speech) system by carefully analyzing a piece of generated speech for content, intonation, prosody, pronunciation, expressiveness, etc.

You are an expert evaluator for voice dialogue systems. Carefully assess the **audio input** based on two dimensions: Speech Quality and Content Quality. 

**Speech Quality**:

\qquad - **Clarity**: Is the speech or audio signal clear and free of noise?
   
\qquad - **Naturalness**: Does the voice resemble a real human without robotic or artificial sounding effects? Is there emotional expressiveness?
   
\qquad - **Continuity**: Are there any interruptions, stutters, or glitches?

**Content Quality**:

\qquad - **Content matching**: Whether the content of the audio transcript solves the key information matching with the reference text. Compare the reference text and the audio transcript.
    
\qquad - **Attribute matching**: Whether the emotion, speed, role and other attributes of the output audio match the expected (consider the history context).

To score, please follow these steps:

1. **Transcription**:

\qquad - First, transcribe the audio as accurately as possible.
   
\qquad - Then use that transcript to evaluate the speech and content quality as described above.

2. **Scoring**:
You will rate each dimension on a scale from **1 to 5**, using the following rubrics:

\qquad - **Speech Quality**:

\qquad \qquad 1: Unintelligible or extremely difficult to understand. Critical flaws: Extremely loud background noise severely impacts comprehension; completely robotic/electronic voice with utterly stiff intonation.
    
\qquad \qquad 2: Sounds like a robot reading a script. Audible but very monotonous voice with zero emotional expression, similar to navigation systems or early AI voices.
    
\qquad \qquad 3: Sounds human but lacks emotion or has noticeable flaws. Clear speech that sounds human, but has flat intonation without emotional variation. May contain noticeable defects (e.g., occasional weird pronunciation, slightly muffled sound).
    
\qquad \qquad 4: Basically indistinguishable from human but not perfect. Clear speech resembling human speaking with basic intonation variations and some emotion. May have very minor flaws that don't affect overall listening experience.
    
\qquad \qquad 5: Perfect and indistinguishable from human. Completely clear voice that not only sounds human but is emotionally rich with proper prosody and modulation, showing full expressiveness.

\qquad - **Content Quality** (Two-Step Method):

\qquad \qquad **Step 1: Determine Category**

\qquad \qquad \qquad - If BOTH content and attributes are poor → Score 1 (End evaluation)
        
\qquad \qquad \qquad - If ONE aspect is good but the other is poor → Proceed to 2-3 score range
        
\qquad \qquad \qquad - If BOTH content and attributes are good → Proceed to 4-5 score range

\qquad \qquad **Step 2: Detailed Scoring**

\qquad \qquad \qquad For 2-3 Range (One aspect deficient):
        
\qquad \qquad \qquad \qquad - Score 2: If attributes are severely mismatched (even if content is acceptable). Examples: Required role play completely missed; requested slow speed but too fast to understand; emotional tone completely wrong.
        
\qquad \qquad \qquad \qquad - Score 3: If attributes are relatively well-matched (regardless of content quality). Examples: Emotional expression mostly appropriate; role play generally convincing despite content issues.

\qquad \qquad \qquad For 4-5 Range (Both aspects good):
        
\qquad \qquad \qquad \qquad - Score 4: Accurately solves the problem and clearly meets attribute requirements. Standard, satisfactory completion.
        
\qquad \qquad \qquad \qquad - Score 5: Perfectly solves the problem (may include extra value) and demonstrates highly precise attribute fulfillment. Examples: Role play is vivid and authentic; instruction following is exceptionally well-executed.

3. **Output Format**:
You must respond with a JSON object in the following format:

\{

\qquad "transcript": "The recognized spoken content", 
  
\qquad "speech\_quality\_score": int (1-5),
  
\qquad "content\_quality\_score": int (1-5),
  
\qquad "speech\_score\_reasoning": "Brief reasoning that explains your speech\_quality\_score, especially highlighting key strengths or issues in speech clarity and expressiveness.",
  
\qquad "content\_score\_reasoning": "Brief reasoning that explains your content\_quality\_score, especially highlighting key strengths or issues in semantic accuracy and alignment with background."
  
\}

You will be provided with:

\qquad - **background\_text**: Provides key context information to judge the synthesized speech.

\qquad - **audio**: The audio generated by the end2end system to be evaluated.

The background text:

\{background\_text\}

And, the synthesized speech from the system, please analyze it carefully

**synthesized\_speech**
\end{mdframed}

\end{document}